\documentclass[journal,10pt]{IEEEtran}

\usepackage{pifont}
\usepackage{lscape}
\usepackage{adjustbox}
\usepackage{cite}

\newcommand{\cmark}{\ding{51}} 
\newcommand{\xmark}{\ding{55}} 
\newcommand{\pmark}{\ding{109}} 

\usepackage[T1]{fontenc}
\usepackage[utf8]{inputenc}
\usepackage[english]{babel}

\usepackage[cmex10]{amsmath} 
\usepackage{amssymb,amsfonts,bm}
\usepackage{amsthm} 

\usepackage{graphicx}

\usepackage{array,booktabs,tabularx}
\usepackage{ragged2e}
\usepackage{multirow}
\usepackage{stfloats}
\newcolumntype{C}[1]{>{\centering\arraybackslash}m{#1}}
\newcolumntype{Y}{>{\RaggedRight\arraybackslash}X}
\newcolumntype{W}[1]{>{\hsize=#1\hsize\RaggedRight\arraybackslash}X} 
\renewcommand{\arraystretch}{1.15} 
\newcolumntype{P}[1]{>{\centering\arraybackslash}p{#1}} %

\usepackage[table,xcdraw]{xcolor}
\usepackage{color,soul}
\soulregister\cite7
\soulregister\ref7

\usepackage{subcaption}

\usepackage{float}
\usepackage{balance}

\usepackage{soul}        
\usepackage{csquotes}
\usepackage{textcomp}
\usepackage{dirtytalk}
\usepackage{wrapfig}

\usepackage[hidelinks]{hyperref} 

\usepackage[ruled,vlined]{algorithm2e}

\usepackage[acronym]{glossaries}
\makeglossaries

\definecolor{headergray}{HTML}{E0E0E0}

\newacronym{gpai}{GPAI}{Generative Physical Artificial Intelligence}
\newacronym{rfm}{RFM}{Robot Foundation Model}
\newacronym{vla}{VLA}{Vision Language Action Model}
\newacronym{lbm}{LBM}{Large Behavior Model}
\newacronym{wfm}{WFM}{World Foundation Model}
\newacronym{dpm}{DPM}{Diffusion Policy Model}

\newacronym{cnn}{CNN}{Convolutional Neural Network}
\newacronym{vit}{ViT}{Vision Transformer}
\newacronym{imu}{IMU}{Inertial Measurement Unit}
\newacronym{llm}{LLM}{Large Language Model}
\newacronym{vlm}{VLM}{Vision–Language Model}
\newacronym{dppo}{DPPO}{Diffusion Proximal Policy Optimization}
\newacronym{hri}{HRI}{Human-Robot Interaction}
\newacronym{dof}{DoF}{Degrees of Freedom}
\newacronym{film}{FiLM}{Feature-wise Linear Modulation}
\newacronym{rtone}{RT-1}{Robotics Transformer 1}
\newacronym{rttwo}{RT-2}{Robotics Transformer 2}
\newacronym{aura}{AuRA}{Autonomous Robot Architecture}
\newacronym{cagr}{CAGR}{Compound Annual Growth Rate}
\newacronym{lidar}{LiDAR}{Light Detection and Ranging}
\newacronym{radar}{RADAR}{Radio Detection and Ranging}
\newacronym{rl}{RL}{Reinforcement Learning}
\newacronym{hrl}{HRL}{Hierarchical Reinforcement Learning}
\newacronym{rlaif}{RLAIF}{Reinforcement Learning from AI Feedback}
\newacronym{ppo}{PPO}{Proximal Policy Optimization}
\newacronym{sac}{SAC}{Soft Actor-Critic}
\newacronym{rlhf}{RLHF}{Reinforcement Learning from Human Feedback}
\newacronym{dt}{DT}{Digital Twin}
\newacronym{adas}{ADAS}{Advanced Driver Assistance System}
\newacronym{sdk}{SDK}{Software Development Kit}
\newacronym{mpc}{MPC}{Model Predictive Control}
\newacronym{use}{USE}{Universal Sentence Encoder}

\begin{document}

\title{A Comprehensive Review of Generative Physical Artificial Intelligence}
    
\author{Satyam Gaba, Krutiksinh Rana, Siva Sai, Vinay Chamola~\IEEEmembership{Senior Member,~IEEE}, Dusit Niyato \IEEEmembership{Fellow,~IEEE}

\thanks{This work was partly supported by the CHANAKYA Fellowship Program of TIH Foundation for IoT \& IoE (TIH-IoT) received by Dr. Vinay Chamola under Project Grant File CFP2022027. }
\thanks{Satyam Gaba is with Qualcomm Research, San Diego, USA, and University of the Cumberlands, KY, USA. The work is done outside of the role at Qualcomm Technologies Inc. (e-mail: sgaba@qti.qualcomm.com)}

\thanks{Krutiksinh Rana is with Department of Electronics and Instrumentation, BITS-Pilani, Goa Campus, India (e-mail: f20231262@goa.bits-pilani.ac.in)}

\thanks{Siva Sai is with the Department of Electrical and Computer Engineering, National University of Singapore, Singapore 119077 (e-mail:
siva.sai@nus.edu.sg).}
\thanks{Vinay Chamola is with the Department of Electrical and Electronics Engineering, BITS-Pilani, Pilani Campus, India 333031 (e-mail:  vinay.chamola@pilani.bits-pilani.ac.in).}
\thanks{Dusit Niyato is with School of Computer Science and Engineering, Nanyang Technological University, Singapore (e-mail:  dniyato@ntu.edu.sg).}

}

\maketitle
\begin{abstract}
The integration of large-scale foundation models with physical embodiments has led to significant advancements in robotics termed \gls{gpai}. These agentic AI systems autonomously perceive, reason, and act in complex real-world situations. This survey comprehensively analyzes \gls{gpai} systems, focusing on their architectural foundations, current applications, and key limitations. We introduce a taxonomy of five distinct approaches: \glspl{rfm} for cross-platform skill transfer; \glspl{vla} for end-to-end multi-modal perception and control; \glspl{lbm} for human-like movement generation; \glspl{dpm} for diffusion model-based temporally coherent action generation; and \glspl{wfm} for physics-compliant simulation and data generation. We examine how these approaches complement each other: WFMs generate training data for VLAs and DPMs, RFMs enable cross-platform deployment of learned policies, while LBMs provide motion priors for natural behavior. Through examples across autonomous vehicles, industrial automation, healthcare robotics, and humanoid systems, we identify significant performance improvements and summarize promising research directions in data-efficient learning, sim-to-real transfer, edge-compatible architectures, and safety frameworks. These insights advance embodied AI for IoT-connected environments where intelligent agents interact with networked sensors, actuators, and edge devices.

\end{abstract}
\section{Introduction}
\label{sec:intro}

\glsresetall

\gls{gpai} refers to the application of large-scale generative models to directly synthesize actions, trajectories, and environment predictions for autonomous physical systems. Unlike classical robot learning, which relies on task-specific architectures trained on narrow datasets, \gls{gpai} leverages foundation models pre-trained on diverse multi-modal data to enable zero-shot generalization across tasks, objects, and embodiments. Rather than mapping observations directly to actions, \gls{gpai} systems iteratively generate and refine control policies through sampling-based processes. By prioritizing scalability and cross-domain transfer over task-specific optimization, \gls{gpai} allows robots to adapt to novel scenarios without extensive retraining. While traditional generative models excel at producing digital content such as text and images, \gls{gpai} extends these capabilities to closed-loop physical control under real-world constraints. This advancement is rapidly redefining the boundaries of autonomy, adaptability, and intelligence in fields such as robotics, autonomous vehicles, manufacturing, and healthcare.

\subsection{Evolution from Traditional Systems}

\subsubsection{Limitations of Early Robotic Systems}

Early systems, such as industrial robots, relied on pre-programmed rules and deterministic algorithms. These systems operated on strict if-then logic for specific tasks, generating scripted motion sequences and sensor-triggered responses. This rigidity often proved ineffective when faced with unexpected scenarios and could only function within narrowly defined parameters. For example, early automotive assembly-line robots followed rigid, pre-programmed paths using simple if-then logic for tasks such as spot welding or pick-and-place, and could not adapt to changes in part position, tool wear, or unexpected obstacles. Modern \gls{gpai}, in contrast, leverages foundation models, large pre-trained neural networks capable of generalizing across tasks, embodiments, and environments. This shift is catalyzed by advances in \glspl{rfm}, \glspl{vla}, \glspl{lbm}, \glspl{wfm} and \glspl{dpm}.

\subsubsection{Traditional Robotic Architectures}

Traditional robotics can be categorized into three fundamental approaches: Hierarchical, Reactive, and Hybrid architecture, based on the interaction and capabilities of the sense, plan, and act modules in a robotic system. \textit{Hierarchical architecture} follows a sequential sense-plan-act cycle with slow response times and poor adaptation to dynamic environments, while \textit{reactive architecture} eliminates planning part for faster responses following sense-act cycle but they lacked reasoning capabilities. \textit{Hybrid architectures} address these shortcomings by combining task planning with reactive execution \cite{gat1998three}, this allows the planning phase to execute independently while maintaining tight sense-act coupling for immediate responses. This PLAN then SENSE-ACT approach enables robots to compute future goals while reactively completing current tasks \cite{gat1998three}.

\subsubsection{Modern GPAI System Architecture}

Modern GPAI systems represent an advanced evolution of hybrid architectures, where planning is replaced by sophisticated generative models such as \glspl{llm} with high-level reasoning and natural language understanding capabilities. The ``sense'' component is enhanced by advancement of \glspl{vlm} that ground abstract reasoning in real-world perception. This evolution greatly improves the system's ability to understand context, adapt to new situations, and learn from experience, creating truly autonomous systems capable of operating in unstructured real-world environments.

\subsection{Current Market Adoption and Industry Impact}

The rise of GPAI has driven significant industrial interest, reflecting both its technical potential and pressing economic drivers such as global labor shortages~\cite{Citi_report}. Market projections estimate that 1.3 billion AI-powered robots could be deployed by 2035, with the humanoid robot sector alone valued at \$38 billion and the broader AI-powered robotics market reaching \$178 billion by 2033~\cite{GoldmanSachs_report, Forbes_JK_article,yang2025global}. Early deployments demonstrate significant impact; for example, Amazon’s integration of over 750{,}000 AI-driven robots has achieved efficiency gains of up to 25\% in some fulfillment centers~\cite{Forbes_JK_article}. While these trends highlight the potential of GPAI, their realization depends on addressing several deployment challenges, including safety certification, regulatory frameworks, integration costs, and workforce adaptation. Taken together, these developments point to a decisive shift in the structure of modern economies, with AI-enabled robotics emerging as a foundational technology shaping how future industries operate~\cite{Deloitte_article}.

\subsection{Related Surveys}

The current survey landscape for \gls{gpai} is growing but remains fragmented, with prior work scattered across robotic manipulation, physics-aware perception, foundation models for robotics, and embodied AI. Several recent studies illustrate the diversity of research in this area. In the context of robotic manipulation, Zhang et al. \cite{zhang2025generative} classified generative models, including GANs, VAEs, and diffusion models, providing a structured overview of how these models support physical control tasks. Liu et al. \cite{liu2025generative} clarified theoretical foundations in this domain by examining physics-aware generative modeling in vision, distinguishing between explicit simulation-based approaches and implicit learning techniques. Xu et al. \cite{xu2024survey} expanded the GPAI landscape by surveying large-scale foundation models, such as \glspl{llm} and \glspl{vlm}, with an emphasis on their roles in perception and planning, as well as their use in benchmarking. In addition, Li et al. \cite{li2025comprehensive} provided a comprehensive taxonomy of world models for embodied AI, analyzing spatial–temporal representation structures and distinguishing between decision-coupled and general-purpose models. Xiao et al. \cite{xiao2025robot} considered how foundation models enable robot learning frameworks to be more scalable by facilitating task learning and skill transfer. However, these studies remain fragmented, each focusing on a specific subdomain without providing an integrated view of embodied-physical intelligence systems as a whole.

Table~\ref{tab:survey-comparison} highlights that existing surveys lack comprehensive coverage across applications, benchmarks, and the full range of GPAI model types required to understand embodied intelligence. Many overlook safety-critical considerations, deployment-oriented evaluation, and standardized benchmarks necessary for scalable implementation. Moreover, no prior framework systematically connects generative techniques across the perception–cognition–actuation pipeline or addresses cross-cutting challenges—such as safety assurance, data scarcity, sim-to-real transfer, and ethics—in an integrated manner.

To address these gaps, this survey provides a comprehensive review of \gls{gpai} through an integrative taxonomy that unifies generative approaches within a consistent framework spanning perception, cognition, and actuation. Our taxonomy encompasses \glspl{rfm}, \glspl{vla}, \glspl{lbm}, \glspl{dpm}, and \glspl{wfm}, enabling direct architectural comparison across deployment contexts. Beyond architectural classification, we incorporate deployment considerations, standardized evaluation benchmarks, and a risk assessment framework covering data limitations, sim-to-real transfer challenges, safety testing, and ethical concerns. By organizing previously scattered research into a cohesive structure, we provide an end-to-end perspective to support more integrated and practical advancements in GPAI and embodied intelligence.

\begin{table*}[tb]
\caption{Comparison of related surveys in Physical AI.}
\label{tab:survey-comparison}
\centering
\small
\setlength{\tabcolsep}{4pt}
\begin{tabularx}{\textwidth}{@{}%
  P{1cm}   
  >{\RaggedRight\arraybackslash}P{2.8cm}   
  Y        
  Y        
  P{1.6cm} 
  P{1.6cm} 
  P{1.6cm} 
@{}}
\toprule
\rowcolor[HTML]{E0E0E0}
\textbf{Ref.} & \textbf{Primary Focus} & \textbf{Key Contributions} & \textbf{Limitations} &
\textbf{Applications} & \textbf{Benchmarks} & \textbf{Coverage} \\
\midrule
'\cite{zhang2025generative} &
Generative models for robotic manipulation &
Introduces hierarchical classification (data/intermediate/policy); details GANs, VAEs, diffusion approaches &
Limited to robotic manipulation; lacks multi-domain perspective or system risk evaluation &
\xmark & \pmark & \pmark \\
\addlinespace
\cite{liu2025generative} &
Physics-aware generative AI in vision &
Paradigm taxonomy for explicit/implicit physics; analyzes evaluation protocols &
Restricts to vision, omits agent embodiment, system context, or risk &
\xmark & \pmark & \xmark \\
\addlinespace
\cite{xu2024survey} &
Foundation models in robotics &
Reviews LLMs/VLMs in perception/planning; covers datasets and benchmarks &
Lacks granular generative model integration, minimal focus on simulation or control &
\cmark & \cmark & \pmark \\
\addlinespace
\cite{li2025comprehensive} &
World models for embodied AI &
Unified three-axis taxonomy; formalizes decision-coupled vs. general-purpose models; spatial/temporal representation analysis &
Limited to world model architectures; minimal coverage of other generative approaches (GANs, VAEs) or risk assessment &
\pmark & \cmark & \pmark \\
\addlinespace
\cite{xiao2025robot} &
Robot learning with foundation models &
Explains evolution of robot learning, task mapping, platform comparisons &
Does not provide explicit generative model or sim2real taxonomy; lacks cohesive risk analysis &
\cmark & \pmark & \xmark \\
\addlinespace
\textbf{Our Paper} &
GPAI: Systematic, cross-domain review &
Integrated taxonomy; synthesizes architecture, cross-sector applications &
Full empirical deployment outside scope; benchmarks are continually evolving &
\cmark & \cmark & \cmark \\
\bottomrule
\end{tabularx}

\vspace{0.4em}
\begin{minipage}{\linewidth}
\footnotesize
\textbf{Legend:}
\textbf{Applications} refers to discussion of real-world domains (e.g., robotics, automation) and their use-cases.
\textbf{Benchmarks} relates to standardized datasets or evaluation protocols.
\textbf{Comprehensive Coverage} indicates whether the survey addresses the full spectrum of GPAI models (e.g., LBMs, VLAs, DPMs, diffusion models, world models). \\
\textbf{Symbols:} \cmark = comprehensive; \pmark = partially ; \xmark = not covered.
\end{minipage}
\end{table*}


\subsection{Research Contribution and Scope}

The main contributions of our work are as follows:
\begin{enumerate}
    \item We present the first comprehensive survey of \gls{gpai} systems, covering architectural foundations, their core components, applications, and deployment reality across robotics and autonomous domains.
    \item We introduce a systematic taxonomy of five GPAI paradigms--\glspl{rfm}, \glspl{vla}, \glspl{lbm}, \glspl{dpm}, and \glspl{wfm}--and highlight their complementary roles in perception–action integration, action policy generation, and data/modeling for world simulation.
    \item We present a concise component view of GPAI and highlight widely used evaluation benchmarks for consistent comparison across methods.
    \item We survey GPAI applications across key sectors: autonomous vehicles/\gls{adas}, industrial robotics and manufacturing, healthcare and medical robotics, humanoids/\gls{hri}, logistics and supply chain, digital infrastructure/smart systems, and consumer/research use cases, highlighting capabilities, system patterns, and persistent gaps.
    \item We analyze limitations and risks spanning data scarcity, bias and ethics, hardware and real-time constraints, simulation-to-real transfer, generalization and reasoning, integration complexity, fine-grained control, robustness and safety, transparency and explainability, energy efficiency, and scaling/computation costs, and we outline mitigation directions.
    \item We outline future directions centered on data-efficient learning, modular/generalizable foundation architectures, edge-amenable computation for real-time control, and rigorous safety protocols and governance for trustworthy deployment.
\end{enumerate}

The rest of the paper is organized as follows. Section~\ref{sec:components} details the core components of GPAI and the end-to-end closed-loop system. Section~\ref{sec:taxonomy} propose a comprehensive taxonomy of each architecture of GPAI, elaborating \glspl{rfm}, \glspl{vla}, \glspl{lbm}, \glspl{dpm}, and \glspl{wfm}, alongside illustrative comparisons and architecture figures. Section~\ref{sec:applications} surveys applications across domains with representative case studies. Section~\ref{sec:challenges} enumerates real-world challenges and risks, and potential remedies. Section~\ref{sec:future_directions} discusses future research directions. Section~\ref{sec:conclusion} concludes the review.

\begin{table}[htbp]
\caption{Major acronyms used in the survey.}
\label{tab:acronyms}
\centering
\small
\renewcommand{\arraystretch}{1.2}
\setlength{\tabcolsep}{8pt} 
\begin{tabularx}{\linewidth}{|l|X|}
\hline
\rowcolor{headergray}
\textbf{Notation} & \textbf{Meaning} \\
\hline
CNN & Convolutional Neural Network \\
DPM & Diffusion Policy Model \\
DPPO & Diffusion Proximal Policy Optimization \\
DT & Digital Twin \\
GPAI & Generative Physical Artificial Intelligence \\
HRI & Human–Robot Interaction \\
HRL & Hierarchical Reinforcement Learning \\
IMU & Inertial Measurement Unit \\
LBM & Large Behavior Model \\
LiDAR & Light Detection and Ranging \\
LLM & Large Language Model \\
PPO & Proximal Policy Optimization \\
RADAR & Radio Detection and Ranging \\
RFM & Robot Foundation Model \\
RL & Reinforcement Learning \\
RLHF & Reinforcement Learning from Human Feedback \\
RT-1 & Robotics Transformer 1 \\
RT-2 & Robotics Transformer 2 \\
SAC & Soft Actor–Critic \\
USE & Universal Sentence Encoder \\
ViT & Vision Transformer \\
VLA & Vision–Language Action \\
VLM & Vision–Language Model \\
WFM & World Foundation Model \\
\hline
\end{tabularx}
\end{table}

\section{Components of GPAI}
\label{sec:components}

\begin{figure*}
    \centering
    \includegraphics[trim=0cm 12.7cm 0cm 1.6cm, clip, width=0.9\linewidth]{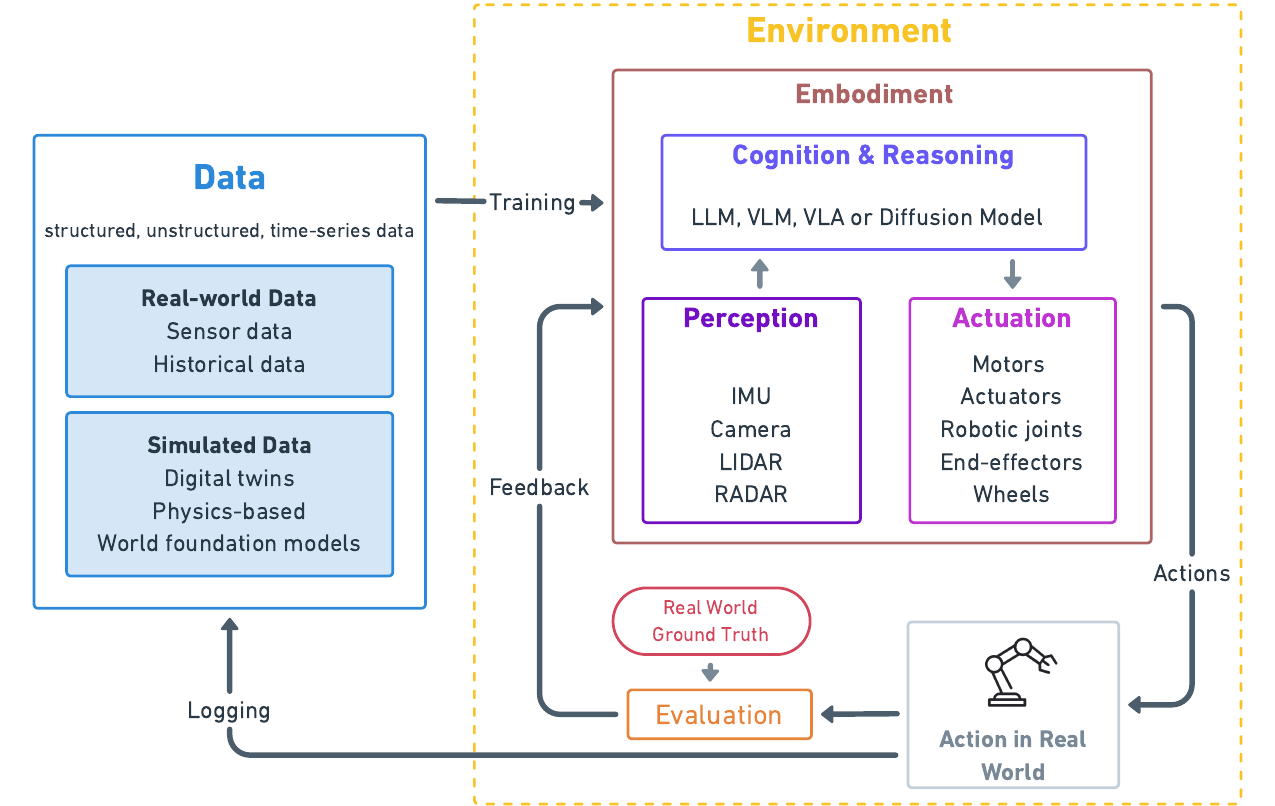}
    \caption{Overview of a GPAI system. Data (real-world or simulated) trains the system and logs embodiment actions. The environment contains the embodiment with perception, cognition \& reasoning, and actuation components. Evaluation compares outcomes against ground truth for closed-loop feedback.}
    \label{fig:components}
\end{figure*}

\subsection{Embodiment}

\gls{gpai} embodiment integrates multiple subsystems into a coherent architecture that operates autonomously in real-world environments. This integration maintains a continuous interaction between cognitive processes and physical reality, enabling adaptive behavior in response to planned objectives and unexpected changes. The system comprises four interconnected subsystems: \emph{cognition} for high-level reasoning and decision-making; \emph{perception} for multimodal environmental understanding; \emph{actuation} for translating decisions into precise motor actions via robotic hardware; and a \emph{continuous feedback loop} that enables real-time adaptation and learning~\cite{sun2024comprehensive}.

As illustrated in the system architecture, these modules reside within the broader environmental context, where continuous feedback from real-world actions flows back into both the perception and data components for evaluation and further refinement. Both perception and actuation subsystems are enhanced through simulation data which provides extensive training scenarios and edge cases, allowing the system to develop robust capabilities before deployment in real-world environments. These embodiment capabilities manifest across diverse physical forms made specifically for various operational domains. Manipulator systems, including robotic arms and grippers, enable precise object handling for manufacturing assembly, warehouse logistics, and healthcare assistance tasks \cite{shademan2016supervised,lykov2024industry}. Mobile robot platforms such as autonomous vehicles for transportation, drones for aerial operations, and ground-based mobile systems are designed for navigation and delivery in complex environments. Humanoid embodiments feature human-like morphology for generalist manipulation tasks and natural human-robot interaction \cite{sheng2025comprehensive}.

\subsection{Cognition \& Reasoning}

\gls{gpai} utilizes complementary strengths of \glspl{llm} and \glspl{vlm}. \gls{llm} provide high-level reasoning and natural language understanding, enabling the system to interpret complex commands and formulate strategic plans. \glspl{vlm} ground these capabilities in real-world perception by combining visual encoders with language models to interpret sensory input such as images. Natural language commands like ``pick up the red cup on the left side of the table'' are processed through a coordinated workflow where the \gls{llm} first parses the command to extract action intent, object descriptors, and spatial references. The \gls{vlm} simultaneously analyzes the scene to identify and localize objects matching these descriptors. The system then performs spatial reasoning to determine the optimal trajectory, considering object properties like size and fragility to select appropriate grip strength. Finally, it generates precise motor commands that translate the high-level understanding into executable physical actions, continuously refining the approach based on real-time visual feedback to ensure successful task completion in dynamic environments. \glspl{vla} provide end-to-end reasoning capabilities that directly translate multimodal understanding into executable action sequences \cite{sapkota2025vision}. They enable seamless integration of vision, language, and action without intermediate processing steps, creating more responsive and contextually aware robotic systems. Diffusion models are also used in cognition. They work by iteratively refining their plans based on real-time sensory feedback, allowing them to adapt to changing environments.

\subsection{Actuation}

In \gls{gpai} systems, once high-level cognitive plans get formulated, they cascade through a control module that transforms abstract intentions into physical motions. This transformation process involves executing precise trajectories that account for workspace constraints, collision avoidance, and optimization criteria such as smoothness and energy efficiency.
The planned trajectories are then processed by inverse kinematics solvers that calculate the specific joint angles and configurations required for each degree of freedom in the robotic system. These calculations must account for the robot's mechanical constraints, joint limits, and singularity avoidance while ensuring the end-effector follows the desired path with appropriate timing and coordination \cite{li2023actuation}.
Low-level real-time controllers convert these computed joint angles into the actual electrical and mechanical commands like position setpoints, velocity profiles, and torque commands that are transmitted to individual actuators such as servo motors, pneumatic cylinders, or hydraulic systems\cite{he2025neurodynamics}. These controllers implement feedback control algorithms that ensure accurate trajectory following despite external disturbances, mechanical compliance, and system dynamics.

\subsection{Perception}
The actuation process operates in a continuous, closed-loop feedback system within the \gls{gpai} embodiment, where multiple streams of sensory information are fed back to inform and adjust ongoing actions within the embodiment. The architecture diagram illustrates how perception modules, incorporating \gls{imu}, camera, \gls{lidar}, and \gls{radar}, enable robust sensing for adaptive operation across various contexts. Visual sensors such as cameras, depth sensors, and \glspl{lidar} give real-time information about object positions, orientations, and environmental changes\cite{shen2024evolutionary}. Proprioceptive encoders monitor joint positions, velocities, and the forces applied to them. Tactile sensors in the end-effector provide crucial information about contact forces, surface properties, and grip stability. This multimodal sensory feedback enables the embodiment to perform real-time adaptations necessary for successful task completion\cite{almujally2024multi}. For example, when holding a delicate object, tactile feedback allows the system to adjust grip strength appropriately to prevent damage, while maintaining a secure hold. Visual feedback enables compensation for object movement or unexpected obstacles, while proprioceptive information helps maintain balance and coordination during intricate manipulation tasks.

The integration of this sensory information with the cognitive and actuation systems creates a truly adaptive autonomous system that can handle the inherent uncertainty and unpredictability of real-world environments, enabling robust performance across diverse tasks and conditions.
Consider a practical example: commanding a robot to make morning coffee. This seemingly simple task requires the robot to process natural language such as ``make coffee,'' understand spatial relationships (e.g., coffee is in the kitchen), navigate using environmental maps while avoiding obstacles, and execute precise manipulation sequences. The robot must perceive machine feedback, which in this case could be indicator lights on the machine, apply an appropriate grip force when handling delicate cups, maintain balance to prevent spills, coordinate brewing timing, and adapt to variations in cup placement and coffee machine interfaces~\cite{mon2025embodied}. Each step demands real-time sensorimotor integration, safety monitoring, and failure recovery capabilities.

\subsection{Data Foundation of \gls{gpai}}

\gls{gpai} models rely on diverse, high-quality, and physically grounded data to perform effectively in real-world environments. Key data types include sensor data such as RGB-D images, audio, force/torque measurements, tactile feedback, proprioception, and environmental readings for comprehensive perception. Demonstration trajectories from human or teleoperated tasks encode expert knowledge through state-action pairs. Synthetic data from simulated environments or \glspl{dt} offers scalable, risk-free training opportunities covering multiple edge cases. Multimodal data, combining text, images, video, and action labels, links semantic understanding with physical interaction for intuitive responses \cite{sliwowski2025reassemble,sai2024pivotal}. This data flows continuously into the system (Fig.~\ref{fig:components}), with logging and feedback mechanisms connecting actions and evaluation results back to the data layer, ensuring continual improvement and dataset enrichment. Real-world datasets are fundamental to \gls{gpai} development, as they capture authentic physical interactions and environmental variability that synthetic data cannot fully replicate. These datasets comprise human demonstrations, teleoperated robot trajectories, and continuous sensor streams from deployed systems, offering the richness of real-world noise and unexpected failures that challenge models to develop robust behaviors \cite{karnan2022socially}. However, the scalability of real-world data collection is limited by human labor requirements, privacy risks in capturing sensitive processes, and intellectual property constraints that restrict dataset sharing, leading to uneven data availability across domains.

Complementing real-world data, simulation-based datasets provide scalable training environments where \gls{gpai} systems can safely explore millions of scenarios without physical constraints. Advanced physics simulators generate photorealistic environments with accurate dynamics, creating synthetic datasets that span extreme conditions and rare events impractical to collect in reality \cite{muratore2022robot}. \glspl{wfm} produce diverse scenarios, including object manipulation tasks, navigation challenges, and human-robot interactions within these virtual environments \cite{agarwal2025cosmos}, enabling systems to learn from millions of simulated experiences before deployment. The generated synthetic data encompasses multimodal observations—RGB images, depth maps, force feedback, and proprioceptive sensor readings—all annotated with corresponding optimal actions and outcomes \cite{agarwal2025cosmos}. This simulation-to-reality transfer allows \gls{gpai} systems to develop robust policies that generalize effectively to real-world scenarios \cite{kar2019meta}.

The data landscape is divided into two primary formats: structured, unstructured, and real-time stream, each serving distinct purposes. \emph{structured datasets} consist of information organized into predefined formats, such as labeled trajectories and annotated datasets, which are ideal for teaching specific, well-defined skills that require precision. \emph{Unstructured datasets} capture raw environmental complexity through sources such as continuous video streams or natural language instructions, enabling systems to learn from real-world ambiguity and complexity. \emph{Real-time data streams} enable continuous adaptation through live sensor feeds and user feedback, creating learning systems that update behaviors based on current context while maintaining core competencies \cite{jiang2024robots}. The foundation of \gls{gpai} lies in data scaling and diversity. Large-scale pretraining on millions of samples across varied tasks, embodiments, and environments ensures that models generalize well. Exposure to diverse physical scenarios—such as various robots, objects, lighting conditions, and weather patterns—fosters robustness and adaptability. Equally important is data quality and fidelity, whether gathered from simulations or real-world interactions, as these are critical for deploying systems in unpredictable settings. Supporting this, a robust data infrastructure is essential for \gls{gpai} development. Tools such as \glspl{dt} enable safe and scalable training and testing by simulating real-world scenarios, rare edge cases, and expanding data diversity and realism. Data management frameworks ensure transparency by detailing dataset characteristics and sources, facilitating structured handling of large data volumes, and enhancing system reliability~\cite{motta2023framework}.

\subsection{Learning Paradigms and Feedback Mechanisms in \gls{gpai}}

\gls{gpai} systems require sophisticated learning mechanisms to develop motor skills and complex behaviors for real-world tasks \cite{zhang2021reinforcement}. Unlike traditional supervised learning approaches that rely on labeled datasets, some \gls{gpai} systems leverage \gls{rl} to acquire manipulation, navigation, and assembly capabilities through trial-and-error experiences, making them particularly well-suited for dynamic and unpredictable physical environments \cite{ibarz2021train}. Building on this foundation, \gls{rl} provides a framework for models to find optimal behaviors through direct interaction with their environment\cite{yuan2024transformer}. The framework consists of an agent that observes the state of its world, selects an action, and receives a numerical reward that represents the desirability of the outcome. The agent's singular goal is to refine its policy, to maximize the total expected rewards \cite{ibarz2021train}. In modern \gls{gpai} systems, deep neural networks are essential for making this process viable. They are capable of interpreting high-dimensional sensory inputs to approximate both the value function and the policy. For continuous tasks, algorithms such as \gls{ppo} \cite{schulman2017proximal} and \gls{sac} \cite{haarnoja2018soft}, are quite effective. These methods are favored for their training stability and their ability to optimally balance exploiting the known good actions with exploring new ones.

Advanced approaches, such as \gls{hrl}, are designed to tackle complex tasks by decomposing them into manageable subtasks with distinct temporal abstractions. It organizes policies into distinct levels, with high-level meta-controllers acting as strategists, and low-level controllers executing fine-grained, specific actions. This division of labor allows the system to plan strategically while acting precisely \cite{kulkarni2016hierarchical}. Furthermore, Human-in-the-Loop learning enhances the \gls{rl} process by integrating human expertise through \gls{rlhf}, where humans provide comparative feedback and demonstrations that guide models toward behaviors aligned with human values \cite{christiano2017deep}. However, human-in-the-loop approaches can also introduce risks such as bias amplification when human feedback mirrors societal prejudices, as well as inconsistent feedback from different human annotators. As a scalable alternative, \gls{rlaif} replaces human preference labels with comparisons generated by a capable auxiliary model, enabling scalable preference optimization when human supervision is expensive or limited \cite{lee2023rlaif}.

In contrast to the exploratory nature of \gls{rl}, supervised learning offers a more direct approach by training models on labeled datasets that map inputs to their corresponding desired outputs. This method excels at learning specific motor skills through demonstration data, like object recognition for manipulation and trajectory planning from expert-demonstrated paths \cite{argall2009survey}. The primary advantage lies in the immediate nature of learning from expert demonstrations, enabling systems to acquire well-defined skills with clear success criteria in structured environments. However, this direct learning approach faces limitations in adaptability in real environments due to its reliance on pre-collected, labeled data. The method struggles with diverse situations that are not seen in training datasets. This approach requires extensive human annotation for each new task or environmental condition, making it most effective for structured physical tasks with predictable outcomes \cite{jha2022imitation}.

Moving beyond the constraints of labeled data, unsupervised learning enables models to find patterns and structures in sensory data without explicit supervision. These systems excel at learning representations of objects, surfaces, and spatial relationships through clustering, dimensionality reduction, and generative modeling techniques \cite{bengio2013representation}. In robotics applications, unsupervised learning helps in the development of world models that capture environmental physics and dynamics, enabling better prediction of action outcomes and more robust decision-making. Where labeled data is unavailable, unsupervised learning enables exploratory behavior and environmental understanding, automatically discovering affordances like the action possibilities offered by objects and surfaces. Such capability allows systems to continuously adapt their understanding as they encounter new physical phenomena, making unsupervised learning especially suitable for unstructured environments where creating labeled datasets is impractical.

Bridging the gap between supervised and unsupervised approaches, self-supervised learning generates supervisory signals from the data itself, eliminating the need for human annotation while maintaining directed learning objectives. \gls{gpai} systems use self-supervised methods to learn from their own environmental interactions, creating prediction tasks that drive meaningful representation learning through techniques such as predicting future sensory states and developing forward models of action consequences \cite{pathak2017curiosity}. This paradigm proves particularly effective for multimodal learning in \gls{gpai}, where systems must integrate information from multiple sensory modalities, including vision, touch, proprioception, and others. Self-supervised learning enables robots to develop rich representations by learning cross-modal predictions, such as predicting the tactile properties of objects (e.g., softness, hardness, texture, compliance) from visual appearance or how they will move when manipulated. The approach scales naturally with interaction data availability, making it ideal for lifelong learning scenarios where \gls{gpai} systems continuously improve their capabilities through ongoing real-world experience \cite{pinto2016supersizing}.

Training \gls{gpai} systems reliably and efficiently requires high-fidelity physics simulation environments such as MuJoCo \cite{todorov2012mujoco}, PyBullet \cite{mower2023ros}, and Isaac Sim \cite{makoviychuk2021isaac}. These simulators provide accurate rigid-body dynamics, realistic friction models, and collision detection, allowing models to learn through thousands of trial-and-error episodes without physical hardware constraints. \glspl{dt}, virtualized counterparts of physical systems maintained through real-time sensor data, enable comprehensive testing and validation before real-world deployment. They provide an effective mechanism to bridge simulation and real-world deployment, with emerging frameworks supporting networked edge architectures\cite{tang2022survey}. The critical challenge of sim-to-real transfer is addressed through domain randomization and progressive training methods that ensure policies remain robust under real-world variability and environmental changes \cite{tobin2017domain}. To ensure safe deployment, trained policies must demonstrate robustness to perturbations, reliability across diverse conditions, and predictable failure modes that can be monitored and constrained during operation.

\subsection{Infrastructure and Evaluation Frameworks}

To measure the potential of these trained models and compare the effectiveness of different learning algorithms, standardized evaluation frameworks are essential. These benchmarks provide consistent metrics and tasks, allowing researchers to gauge progress across the field. Key examples include LIBERO \cite{liu2023libero}, which features 130 tasks designed to test lifelong robot learning and the ability of policy architectures to continuously adapt. RLBench \cite{james2020rlbench}, offering 100 vision-guided manipulation tasks, aids reinforcement and few-shot learning. Open X Embodiment \cite{o2024open}, with over 1 million robot trajectories across 22 embodiments, facilitates cross-embodiment learning with pre-trained model checkpoints. However, recent work demonstrates weak correlation between benchmark metrics and real-world robustness, with generalization failures occurring despite high controlled-environment success rates~\cite{sedlacek2025realm}. Safety-critical aspects, like failure prediction accuracy and human safety margins, remain underrepresented~\cite{tang2024defining}.

\section{Taxonomy of Generative Physical AI}
\label{sec:taxonomy}

\begin{figure*}[th]
    \centering
    \includegraphics[trim=0cm 18cm 0cm 3.3cm, clip, width=\linewidth]{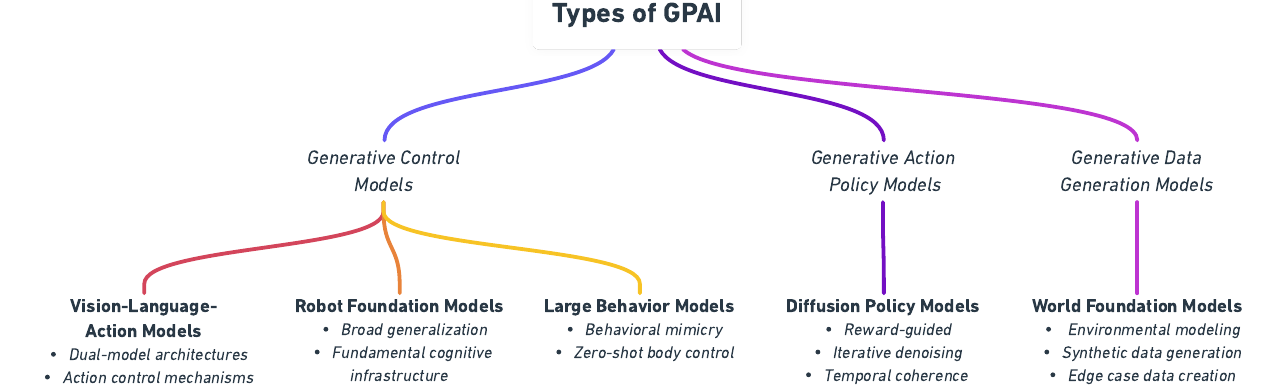}
    \caption{Taxonomy of GPAI showing major model families and their functional roles across control, policy generation, and data synthesis.}
    \label{fig:taxonomy}
\end{figure*}

The taxonomy of \gls{gpai} is organized by functional role: models that integrate perception and control, models that generate and refine action policies, and models that synthesize data for training and evaluation. This classification, illustrated in Fig.~\ref{fig:taxonomy}, captures the complementary pathways through which \gls{gpai} systems achieve reliable physical intelligence.

\emph{Generative Control Models} link sensory input with actuator control commands. \gls{vla}~\cite{brohan2022rt} models align multimodal perception with control heads, enabling translation of natural language or visual prompts into executable actions. \glspl{rfm}~\cite{firoozi2025foundation} extend these capacities through task generalization, providing a cognitive substrate adaptable to diverse robotic embodiments. \glspl{lbm}~\cite{tirinzoni2025zero} emphasizes imitation and zero-shot whole-body control, yielding human-like interaction without task-specific training. \emph{Generative Action Policy Models} specialize in trajectory generation. Diffusion-based approaches~\cite{chi2023diffusion} exemplify this class, employing reward-guided refinement, progressive denoising, and temporal coherence mechanisms to produce stable yet adaptive action sequences. These models strike a balance between responsiveness to dynamic environments and long-horizon consistency. \emph{Generative Data Models} address the demand for scalable and diverse training resources. \glspl{wfm}~\cite{agarwal2025cosmos} simulate high-fidelity environments, create synthetic data at scale, and generate rare edge cases for robustness testing. By expanding available data distributions, \glspl{wfm} supports generalization and safety validation. Taken together, these paradigms underscore the interdependence of \gls{gpai} architectures. Practical systems typically combine perception–action integration from control models, refinement through policy models, and synthetic environments from data models to achieve robust real-world operation.

\subsection{Robot Foundation Models (\glspl{rfm})}

\glspl{rfm} are large-scale, general-purpose AI systems designed to acquire and execute robotic competencies through extensive training on robotics-focused datasets. Unlike conventional models, which are limited to narrow and specific tasks, \glspl{rfm} are trained on vast collections of state–action pairs gathered from diverse robotic platforms. This training paradigm enables them to perform a broad spectrum of robotic tasks with minimal task-specific engineering. \glspl{rfm} embodies the principle of transferring generalizable knowledge across embodiments and tasks, positioning them as a central component in the evolution of generative physical intelligence.

\subsubsection{Multimodal Architecture Framework}

\begin{figure*}[ht]
    \centering
    \includegraphics[trim=0.1cm 13.9cm 0.1cm 4.2cm, clip, width=\linewidth]{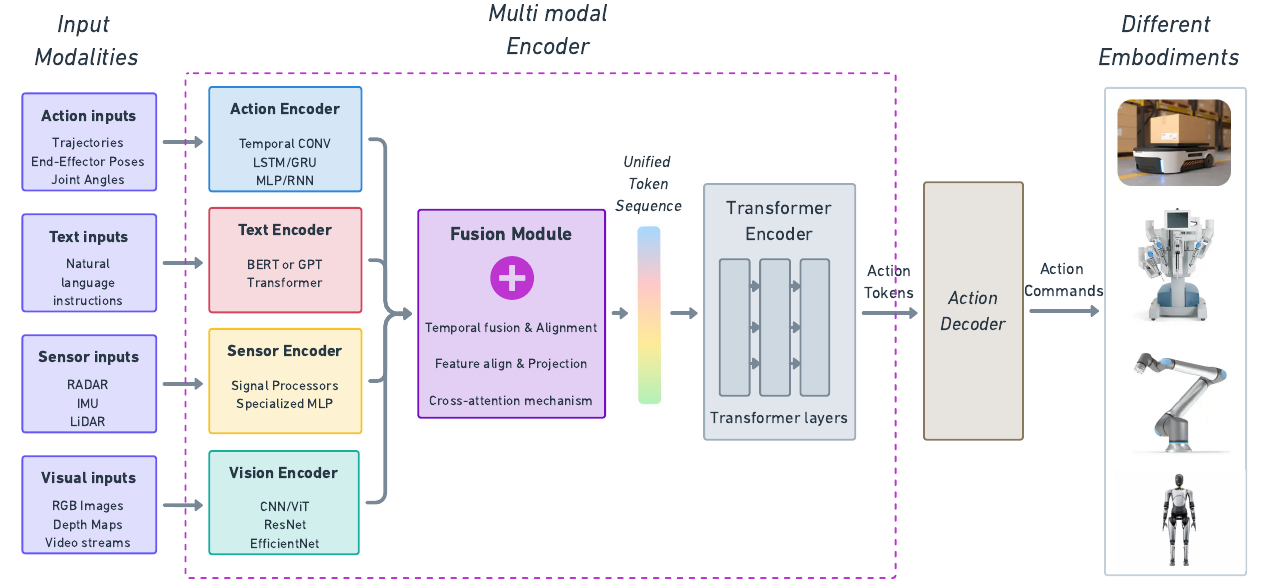}
    \caption{RFM architecture with multimodal encoder system processing action inputs, natural language instructions, sensor data, and visual inputs through specialized encoders. The fusion encoder unifies modalities into token sequences by cross-modal attention in transformer layers. The token sequences pass through the action decoder to produce action commands across diverse robotic embodiments.}
    \label{fig:rfm}
\end{figure*}

The architectural foundation of \glspl{rfm}, as illustrated in Fig.~\ref{fig:rfm}, relies upon multimodal encoder system that concurrently processes diverse input streams, encompassing: (1) action inputs containing trajectory logs and end-effector poses, (2) textual inputs providing natural language instructions for behavioral conditioning, (3) sensor inputs delivering force/torque feedback and proprioceptive data, and (4) visual inputs comprising RGB imagery and depth maps. Each modality undergoes processing through specialized encoders, including temporal convolutions for action sequence analysis, BERT or GPT-based transformer architectures for language comprehension, signal processors for sensor data interpretation, and \glspl{vit} or \glspl{cnn} architectures for visual perception~\cite{devlin2019bert}. These processed modalities are subsequently unified through a fusion encoder that generates a coherent token sequence representing the complete contextual information. This unified token representation is then processed through transformer layers that facilitate cross-modal attention mechanisms, enabling the model to comprehend complex relationships between visual observations, textual commands, and physical feedback signals. The model subsequently produces executable actions through an action decoder capable of controlling diverse embodiments ranging from autonomous vehicles to robotic manipulators to humanoid systems. This architectural paradigm enables \glspl{rfm} to achieve remarkable generalization capabilities across diverse robotics tasks by learning unified representations from multimodal robotics data, thereby allowing a single model to understand and execute complex robotic behaviors across different platforms and operational scenarios.

\subsubsection{Model Categories and Approaches}

\glspl{rfm} can be categorized into two major approaches that differ in scope and functionality: task-specific models and general-purpose models. Task-specific models, which dominated early robotic research, are optimized for narrow domains such as perception or grasp planning. These models achieve high accuracy in repetitive tasks under constrained conditions but lack adaptability to novel situations. They often require manual reconfiguration or retraining when applied to new tasks, which limits their scalability despite having lower computational demands and minor data requirements. An example is GraspClutter6D~\cite{back2025graspclutter6d}, a system engineered exclusively for robotic working in cluttered settings such as warehouses and bins, where it demonstrates state-of-the-art performance in predicting effective grasp strategies.

In contrast, general-purpose models integrate perception, planning, and control within a single representational framework, enabling them to operate across various robotic domains \cite{wang2025unified}. These systems achieve high adaptability and generalizability, though at the cost of significantly greater computational resources and large-scale datasets. They are designed for deployment across heterogeneous robotic platforms, from autonomous vehicles to humanoids, by leveraging unified sensor–actuator representations. A prominent example is PaLM-E~\cite{driess2023palm}, a multimodal foundation model capable of reasoning and planning across language, vision, and robotics tasks. PaLM-E can seamlessly transfer across embodiments and task domains without explicit retraining, demonstrating the scalability of the general-purpose approach.

\subsubsection{Exemplary General-Purpose Models}

Several models have emerged as representative examples of this general-purpose approach, each demonstrating unique methodological strategies for achieving unified functionality across diverse robotics applications. PaLM-E represents a language-centric architectural approach that extends the capabilities of a pre-trained \gls{llm} by incorporating continuous embodied observations directly into its embedding space. The model employs a decoder-only transformer structure based on Google's PaLM language model, enhanced with modality-specific encoders that transform images, robot states, and sensor data into embeddings compatible with the language model's dimensional space. PaLM-E processes "multimodal sentences" where text tokens are strategically interleaved with encoded observations, enabling autoregressive text generation while facilitating reasoning across visual and embodied modalities. The architecture preserves inherent language capabilities while utilizing specialized encoders such as \glspl{vit} for image processing and proprioceptive encoders for robot state estimation, thereby enabling the system to generate step-by-step plans and adapt to environmental changes through natural language outputs that require external interpretation for action execution \cite{zitkovich2023rt}.

Gato~\cite{reed2022generalist}, in contrast, implements a fundamentally different sequence modeling approach wherein every possible input and output, regardless of modality, is tokenized into a unified vocabulary space processed by a single transformer network. Universal tokenization strategies convert heterogeneous data types into discrete tokens. Tokenizers transform text into subwords, ResNet encoders process images into 16×16 patch sequences, discretizers bin continuous actions into predetermined categories, and serializers convert sensor observations into structured token sequences. Gato's architecture treats all modalities equivalently within a single parameter space, utilizing autoregressive cross-entropy loss across hundreds of tasks simultaneously. This design paradigm enables the model to directly generate action tokens, text completions, or visual predictions without requiring architectural modifications. The approach eliminates modality-specific encoders, instead relying on the transformer's self-attention mechanisms to learn cross-modal representations naturally through sequence processing. This methodology enables seamless transitions between conversational dialogue, image captioning, and direct robotic control within a single, unified network architecture. \glspl{llm} like GPT-4, while not initially designed for robotics applications, demonstrate exceptional high-level planning and reasoning capabilities when integrated with physical robotic systems \cite{o2025exploring}. The model excels at interpreting complex natural language commands, decomposing them into actionable step sequences, and generating executable code for robotic systems. This positioning makes it an effective cognitive engine for robotic platforms when paired with appropriate execution frameworks and robotic middleware systems.

\subsection{Vision Language Action (VLA) Models}

\subsubsection{Defining VLA Architecture}
\glspl{vla} constitute a class of multimodal \gls{gpai} systems that tightly integrate visual perception, natural language understanding, and robotic action generation within a unified framework. Unlike general-purpose \glspl{rfm}, \glspl{vla} are designed specifically for embodied AI applications, incorporating architectural elements optimized for sensorimotor control. A defining feature of VLAs is their dual-model structure \cite{hao2025visual}, which combines \glspl{vlm} components for multimodal reasoning with action-generation modules implemented through transformer decoders or diffusion policies. These systems utilize specialized visual encoders, semantic projectors, and language models to achieve direct, end-to-end mapping from multimodal inputs to motor commands, representing a significant advancement in the design of embodied AI systems.

\subsubsection{Multimodal Processing Pipeline}

\begin{figure*}[th]
    \centering
    \includegraphics[trim=0cm 13.5cm 1cm 2.45cm, clip, width=0.8\linewidth]{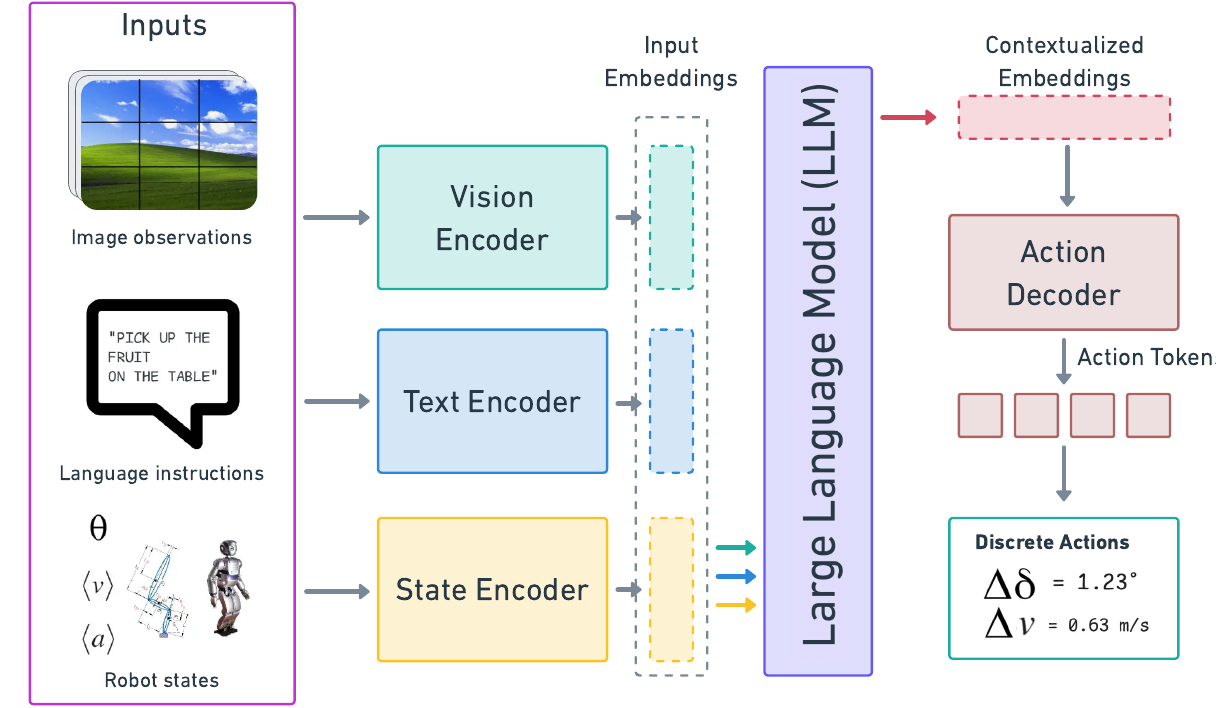}
    \caption{General architecture of a VLA. Image, language, and proprioceptive inputs are separately encoded into embeddings, fused by an LLM into contextualized multimodal embeddings, and then decoded into action tokens by the Action Decoder for robot control.}
    \label{fig:vla}
\end{figure*}

The multimodal processing pipeline of VLA models, illustrated in Fig.~\ref{fig:vla}, begins with three primary input modalities: image observations, natural language instructions, and proprioceptive state information. Image observations are processed through \gls{vit}, which tokenizes image patches into embeddings that capture spatial–temporal environmental features. Natural language instructions are tokenized using models such as \gls{use}~\cite{cer2018universal}, producing semantic embeddings for conditioning downstream action policies. Proprioceptive state encoders~\cite{kim2024openvla} convert joint angles, velocities, and force measurements into temporally consistent tokens. These three input streams, visual, linguistic, and proprioceptive, are then projected into a shared embedding space and fed as discrete token sequences into the \gls{llm} core. Within the LLM, multimodal attention layers operate across the concatenated sequences, enabling cross-modal fusion: visual tokens can attend to language and proprioceptive tokens, and vice versa. This fusion process enables the model to jointly reason over heterogeneous inputs, rather than treating each modality in isolation, and yields a unified latent representation that integrates perception, instruction, and state. The fused representation is subsequently passed through an action decoder, a specialized module that translates the unified multimodal embedding into a sequence of action tokens. These tokens can represent either low-level motor commands, such as joint torques or velocities, or higher-level policy primitives, such as grasp, move, or release. The action decoder is trained to align the multimodal latent space with the robot’s control interface, ensuring that actions are temporally consistent and physically executable. Unlike RFMs that are adapted for robotics tasks through auxiliary control heads or external policy layers added post-training, VLAs integrate the action decoder into their core architecture and train it jointly with the visual and language components. This end-to-end training approach enables gradients from the robot's control objectives to flow back through the entire multimodal system, allowing the visual encoders, language model, and action decoder to co-adapt for sensorimotor control, rather than treating action generation as a separate post-processing step.

\subsubsection{RT-1: Pioneering Transformer-Based Robotics}

\gls{rtone}~\cite{brohan2022rt} marked a pioneering milestone in applying transformer architectures to large-scale robotics, demonstrating the feasibility of training generalizable manipulation policies from extensive datasets of teleoperated demonstrations. The model processes natural language instructions and visual observations to generate sequential robotic actions through a decoder-only architecture built upon a 35-million-parameter transformer backbone. In this framework, natural language instructions are embedded using \gls{use}~\cite{cer2018universal}, while robot camera inputs are processed through EfficientNet-B3~\cite{tan2019efficientnet} to generate visual tokens. These multimodal embeddings are further conditioned through Feature-wise Linear Modulation (FiLM) layers~\cite{perez2018film}, which enable dynamic adaptation of vision features in response to linguistic input, thereby aligning perception and language for grounded action generation. Central to the architecture is the TokenLearner module, positioned between the vision encoder and the transformer core. This component strategically reduces computational overhead by compressing the sequence of visual tokens before multimodal fusion. Specifically, TokenLearner applies attention-based mechanisms with learnable importance weights to identify spatially salient features, reducing the representation from 81 visual tokens to 8 critical tokens. This efficiency mechanism maintains performance while improving scalability by minimizing the computational burden of downstream processing. The compressed tokens are then fused with language embeddings and passed through the transformer layers for sequential reasoning.

The action generation process is performed by the Action Decoder, which translates the transformer’s multimodal outputs into executable control commands. Actions are represented in a discretized format, where each of the 11 action dimensions is quantized into 256 bins. This tokenized representation encompasses positional displacements ($\Delta$Pos X, Y, Z), rotational adjustments ($\Delta$Rot X, Y, Z), gripper state transitions, and termination signals. Such a design enables precise six-degree-of-freedom (6-DoF) manipulation and grasping capabilities while ensuring computational tractability. By leveraging this structured tokenization, \gls{rtone} supports fine-grained robotic control across diverse manipulation scenarios. Experimental evaluations highlight the effectiveness of RT-1, demonstrating a 67\% success rate across more than 700 distinct manipulation tasks. These results underscore the model’s ability to generalize beyond individual training demonstrations, establishing \gls{rtone} as a foundational system that bridges large-scale multimodal learning with practical robotic execution.

\subsubsection{RT-2: Actions as Language}

\gls{rttwo}~\cite{zitkovich2023rt} advances the transformer-based robotics paradigm by conceptualizing robotic actions as a language within a unified representational framework, enabling direct translation of semantic knowledge into executable behaviors. This formulation bridges high-level reasoning and low-level control, replacing modular system architectures with a single integrated pipeline. By leveraging pre-trained vision–language representations, \gls{rttwo} minimizes the requirement for robot-specific data while scaling model capacity from \gls{rtone}’s 35M parameters to 55B-parameter foundation architectures. The system integrates large-scale, pre-trained components, including PaLI-X for visual interpretation and object grounding, and PaLM-E for language and sensor fusion, which are trained jointly through a co-fine-tuning strategy that combines robotics-specific datasets with web-scale multimodal corpora. This approach preserves broad reasoning ability while progressively aligning the model with embodied skills. The defining feature of \gls{rttwo} is its unified token representation, where robotic actions are embedded as text-like tokens within the same representational space as words and visual tokens. In this scheme, the model “speaks” actions as an extension of natural language, allowing for a straightforward yet powerful mechanism to align semantic understanding with control execution. This design enables \gls{rttwo} to inherit knowledge directly from large-scale web data, translating visual recognition and linguistic reasoning into executable action sequences with minimal fine-tuning. The result is enhanced generalization to novel tasks and environments, improved reasoning about object–action relationships, and scalable transfer of multimodal knowledge to embodied domains. \gls{rttwo} thus represents a decisive progression toward general-purpose robotic systems, demonstrating how large-scale language and vision priors can be directly operationalized into robotic action \cite{zitkovich2023rt}.

\subsubsection{Other Contemporary VLA Implementations}

Several contemporary models demonstrate alternative approaches to integrating vision, language, and action within robotics. OpenVLA~\cite{kim2024openvla} provides an open-source framework that combines a Prismatic-7B vision–language backbone with Llama 2, targeting robotic manipulation tasks with strong zero-shot generalization to novel objects and environments. The system incorporates a dual visual encoder that fuses DINOv2 and SigLIP features, mapping continuous robot actions into discrete tokens aligned with the language model’s vocabulary space. This design enables natural language instruction following while supporting precise robotic control. Notably, OpenVLA demonstrates robust performance in end-effector manipulation tasks, surpassing larger closed-source models such as RT-2-X across 29 benchmarks while operating with seven times fewer parameters. The model discretizes seven-degree-of-freedom action spaces into 256 bins, reserving dedicated action tokens within the Llama tokenizer’s vocabulary to ensure alignment between language and action domains.

In contrast, Gemini Robotics VLA integrates Google’s Gemini 2.0 multimodal capabilities with the ALOHA teleoperation platform to enable complex bimanual manipulation tasks \cite{team2025gemini}. Leveraging advanced vision–language modeling, Gemini Robotics VLA demonstrates coordinated dual-arm behaviors that extend beyond traditional single-arm manipulation, highlighting the potential of embodied AI systems for dexterous, human-like physical interaction. Whereas OpenVLA emphasizes accessibility, parameter efficiency, and broad generalization across diverse robotic embodiments through fine-tuning strategies, Gemini Robotics VLA prioritizes high-fidelity control for demanding scenarios requiring synchronized bimanual coordination. This distinction emphasizes the diversity of approaches emerging within the VLA paradigm. Collectively, these models exemplify the frontier of embodied AI research, showcasing how vision–language–action architectures enable robots to interpret and execute complex multimodal instructions with flexibility. Their demonstrated capacity for task and environmental generalization positions VLAs as foundational systems for applications spanning domestic robotics, industrial automation, and collaborative human–robot interaction.

\subsection{Large Behavior Models (LBMs)}

\subsubsection{Definition and Learning Mechanisms}

\glspl{lbm} are computational systems designed to learn, reproduce, and generate complex sequences of physical actions, extending the paradigm of LLMs that focus on text and dialogue. LBMs are trained on multimodal datasets, including text, images, video, and sensor streams, to capture and synthesize human-like behaviors and interactions with physical objects. A central mechanism in LBMs is \gls{rl}, which enables dynamic adaptation to feedback from real-world environments. Unlike static dataset training, reinforcement-driven learning allows LBMs to adjust their behavior in real-time to accommodate uncertainty and change. This adaptive capacity is critical for deployment in robotics, embodied assistance, and interactive coaching systems, where effective operation depends on physical engagement and situational awareness.

\subsubsection{Architecture and Learning Framework}

\begin{figure*}[h]
\centering
\includegraphics[trim=0cm 7.8cm 0cm 4.75cm, clip, width=0.9\linewidth]{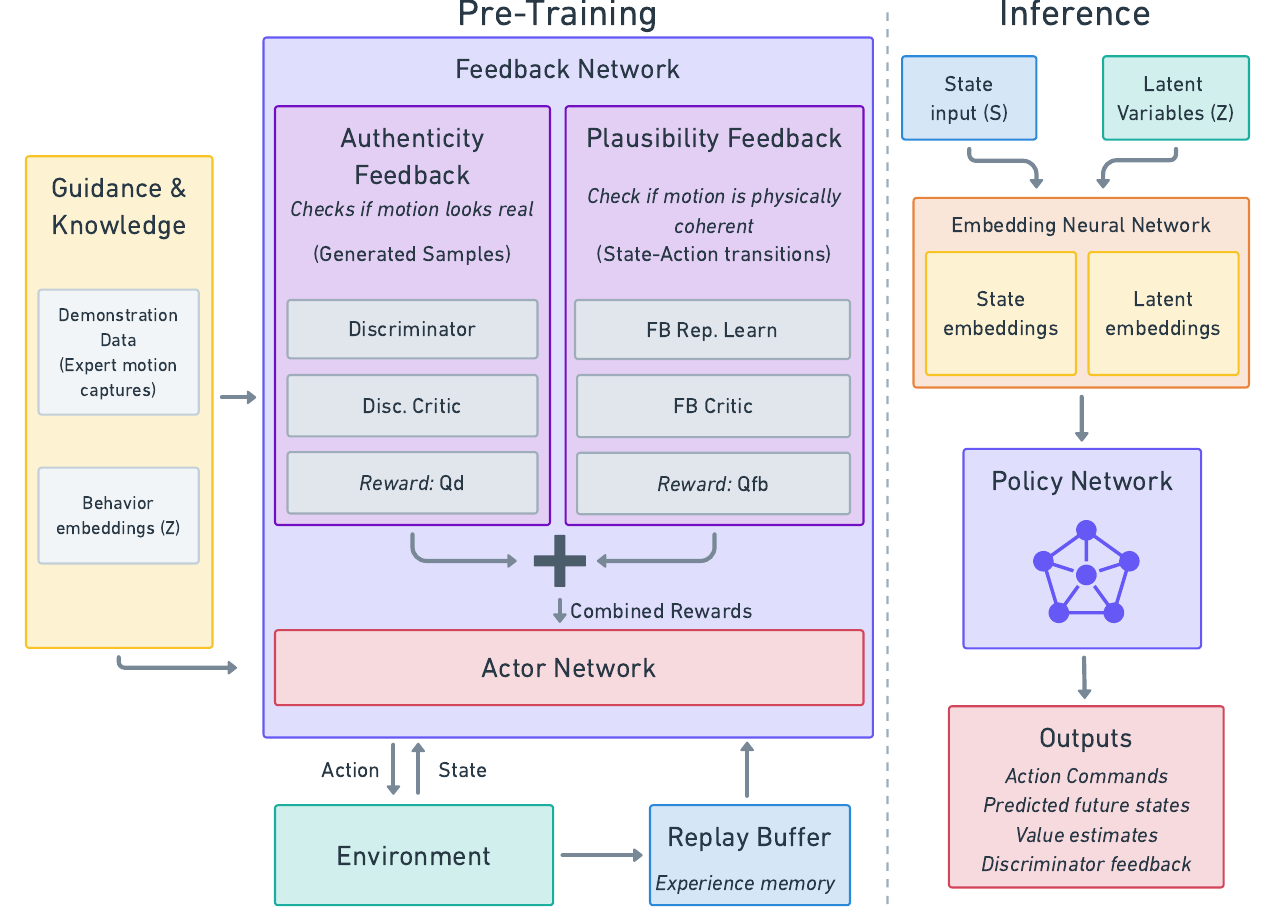}
\caption{Large Behavior Model (LBM) architecture comprising two phases: (left) pre-training with reinforcement learning guided by dual feedback mechanisms (authenticity via discriminator and plausibility via physics-coherence evaluation) on expert demonstration data, and (right) inference with embedding and policy networks for policy-driven real-time action generation from state inputs and latent variables.}
\label{fig:lbm}
\end{figure*}

Fig.~\ref{fig:lbm} illustrates the architecture and workflow of LBMs, emphasizing distinctions from LLMs. In the Pretraining phase, expert demonstration data and behavior embeddings guide the Actor Network, which interacts with simulated or real environments through state-action generation. A Feedback Network supplies two reward signals: authenticity feedback, via a discriminator and critic that evaluate behavioral realism, and plausibility feedback, which assesses the physical coherence of state-action transitions. These combined rewards refine actor policy learning, while a Replay Buffer stores experience for iterative updates. After pretraining, the model enters the Inference phase. State inputs are encoded through an Embedding Neural Network to produce state and latent behavior embeddings. The Policy Network, informed by these representations and latent variables, generates outputs such as action commands, predicted states, value estimates, and discriminator feedback. This design enables LBMs to adapt their behaviors in real-time, update policies through ongoing interaction, and emphasize real-time behavioral grounding, multimodal integration, and continuous adaptation. Their capacity to replicate and refine human-like physical behavior makes them valuable for robotics and embodied AI. Nonetheless, deployment raises risks of misinterpreted actions, unintended behaviors, and physical hazards, necessitating rigorous safety mechanisms and human oversight \cite{khandelwal2023large}.

\subsubsection{Meta Motivo: Zero-Shot Embodied Control}

Meta Motivo exemplifies a \gls{lbm} capable of zero-shot whole-body control without task-specific training. Built on the Forward Backward Representations with Conditional Policy Regularization (FBCPR) algorithm \cite{tirinzoni2025zero}, it converts state observations into humanoid actions via a coordinated multi-network architecture designed to enhance control and generate human-like motion. The system integrates five specialized networks: the Forward Network predicts future states from current state, proposed action, and latent variables; the Actor Network outputs actions as Gaussian distribution means to enable smooth motion; the Critic Network evaluates state–action–latent combinations to provide RL feedback; the Backward Network maps states to latent representations for robust learning; and the Discriminator Network enforces biological plausibility by comparing outputs against unlabeled motion datasets. These foundation models demonstrate advanced behavior generation, yet precise mechanisms remain necessary to translate high-level behavioral intentions into executable robotic commands. This requirement motivates the exploration of specialized methods for action generation.

\subsection{Diffusion Policy Models (DPM)}

\begin{figure*}[h]
    \centering
    \includegraphics[trim=0cm 16.5cm 0.5cm 2.15cm, clip, width=0.9\linewidth]{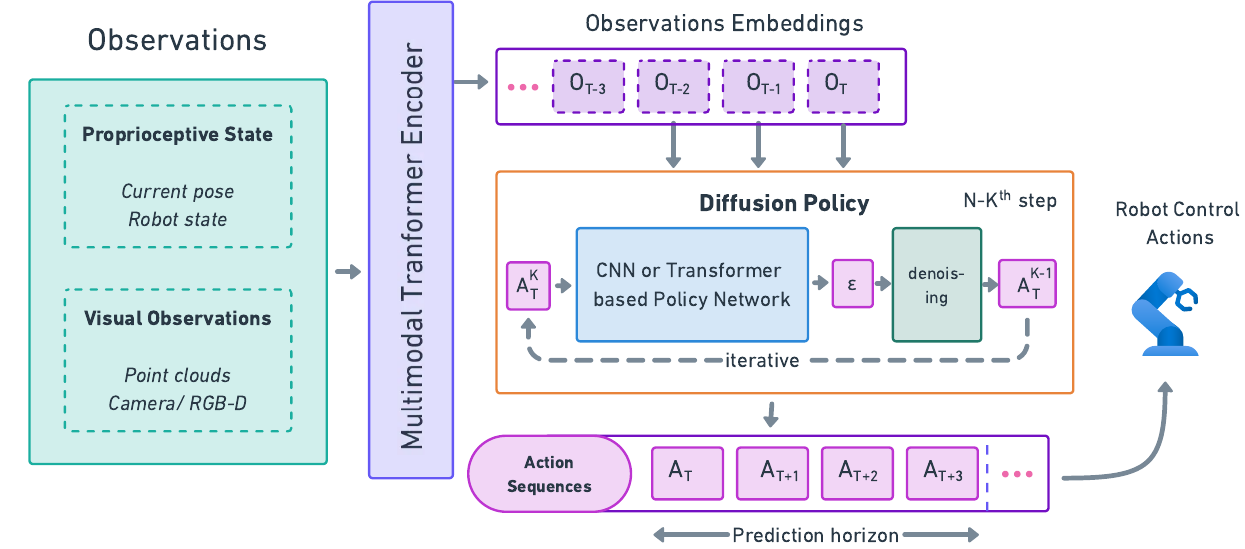}
    \caption{Architecture of Diffusion Policy Models. The observations are processed through a multimodal transformer encoder to generate observation embeddings ($O_{t}$). These embeddings condition an iterative denoising model that predicts noise ($\epsilon$) and refines initially noisy actions ($A_t^k$) to generate coherent action sequences ($A_{t}$) over a prediction horizon for robot control.}
    \label{fig:dpm}
\end{figure*}

\subsubsection{Iterative Action Refinement Framework}

\glspl{dpm} represents a diffusion-based generative framework for robotic control, employing iterative denoising diffusion processes to progressively transform Gaussian noise into coherent, high-dimensional action sequences rather than directly predicting actions\cite{chi2023diffusion}. As illustrated in Fig.~\ref{fig:dpm}, the process begins with multimodal observations, which include proprioceptive data, such as the robot's current pose and state, as well as visual data from point clouds and RGB-D camera streams. This raw sensory input is fed into a multimodal transformer encoder, which generates a sequence of observation embedding over a time horizon \cite{dasari2025ingredients}. These embedding conditions the core of the model, the diffusion policy. The policy starts with a randomly sampled noisy action, denoted as $A_T^K$ in Fig.~\ref{fig:dpm}. This noisy action is iteratively refined in an $N-K$ step denoising loop. In each step, a CNN or Transformer-based policy network predicts the noise ($\epsilon$) present in the current action candidate. This prediction is used to denoise the action, producing an action sequence ($A_T, A_{T+1},\dots$) that is generated over a prediction horizon. The final sequence is then translated into executable robot control actions.

A key strength of this approach is its ability to generate temporally coherent action sequences that extend over 4–8 future steps, enabling stable decision-making across extended horizons~\cite{dasari2025ingredients}. This design directly addresses the requirements of continuous robotic control, in contrast to single-step prediction methods that often struggle with long-horizon stability. Temporal context is maintained by conditioning on stacked observations, preserving relevant historical information. Context-awareness is further enhanced through FiLM layers~\cite{perez2018film} and cross-attention mechanisms, which integrate historical camera frames and proprioceptive signals to dynamically adapt policies to complex sensory conditions. FiLM conditioning applies channel-wide modulation throughout the denoising network, allowing both visual observations and text-based goals to influence attention weights across all layers. This layered conditioning enables policies that are robust to environmental variability and adaptable to real-world multimodal inputs \cite{hou2024diffusion}.

\subsubsection{Model Architecture and Advantages}

\glspl{dpm} leverage receding-horizon control, in which extended action sequences are predicted, the initial steps executed, and plans recalculated frequently to mitigate error accumulation and adapt to dynamic environments. Their core architecture is typically a U-Net-based denoising network that conditions generation on current observations while iteratively refining outputs at each time step. As illustrated in Fig.~\ref{fig:dpm}, the process begins with random Gaussian noise ($A_T^K$) and progressively produces refined actions over $K$ denoising iterations, applying the learned noise-prediction network at each stage. Training involves learning to invert a forward noising process applied to expert demonstration trajectories, with objectives such as L2 loss or epsilon prediction loss used to maximize denoising accuracy. This training paradigm enables the policy to capture the multimodal structure of human demonstrations while maintaining robustness against noise and data imperfections \cite{chi2023diffusion}. Unlike traditional behavior cloning, where deterministic networks average across demonstrations and may yield compromised or invalid actions in multimodal settings, Diffusion Policy models inherently represent multiple valid solutions to complex tasks, thereby avoiding the mode collapse that affects approaches such as LSTM-GMM and Imperical Behavioral Cloning(IBC)~\cite{florence2022implicit}. This property is especially critical in robotics, where manipulation, navigation, and grasping tasks often admit diverse strategies for success. The stochastic iterative denoising process ensures that each generated action is contextually grounded in the current sensory input, producing behaviors that remain consistent with real-world conditions. Empirical evaluations report an average success-rate improvement of 47\% across 12 robotic tasks in 4 benchmark suites, underscoring the advantages of this approach in achieving robust and generalizable control \cite{chi2023diffusion}.

\subsubsection{Training and Evaluation}

Diffusion Policy models are trained on robotic manipulation datasets that contain diverse demonstration data across multiple task domains and complexity levels. Standard Benchmark datasets include RoboMimic \cite{mandlekar2021matters}, which serves as a foundational evaluation benchmark with comprehensive collections of human-teleoperated trajectories across fundamental manipulation tasks such as object lifting, placement, insertion, and multi-arm coordination tasks with varying operator proficiency levels to capture realistic diversity. Multi-environment datasets, including CALVIN \cite{mees2022calvin}, provide recordings across multiple distinct environments for long-horizon language-conditioned tasks. In contrast, datasets such as LIBERO \cite{liu2023libero} contribute diverse manipulation task collections across various evaluation suites, including spatial reasoning, object manipulation, and goal-oriented tasks. 

Specialized metrics are used to assess the practical deployment capabilities of diffusion policy models. Task success rate serves as the primary performance indicator, measuring the percentage of completed manipulation tasks across diverse robotic scenarios and environmental conditions~\cite{chi2023diffusion}. Multimodal action distribution coverage evaluates the model's ability to represent multiple valid solution strategies for complex tasks, thereby mitigating the averaging effects that compromise traditional behavior cloning approaches. Temporal consistency metrics evaluate the coherence and stability of generated action sequences over extended time horizons, which is crucial for long-duration manipulation tasks that require sustained coordination. Real-time inference performance represents another critical evaluation dimension, measuring computational efficiency and response times to ensure compatibility with robotic control systems that operate under strict timing constraints. Cross-platform generalization evaluates model performance across different robot configurations, action spaces (2DoF to 6DoF), and environmental conditions to assess deployment scalability. Robustness under perturbation measures stability when facing observation noise, environmental changes, and unexpected obstacles during task execution.

Evaluation protocols employ systematic testing methodologies across multiple benchmark environments. The assessment framework includes object positioning tasks that require precise spatial alignment, such as Push-T environments~\cite{chi2023diffusion}, which test geometric precision through T-shaped block manipulation with exact target alignment requirements. Multimodal manipulation evaluations assess the model's ability to handle diverse solution strategies through statistical analysis of action distribution coverage and mode preservation. Sequential task evaluations measure long-horizon reasoning capabilities by tracking temporal consistency metrics across extended action sequences~\cite{dasari2025ingredients}. At the same time, household environment assessments featuring interactive components evaluate multi-stage manipulation skills through denoising efficiency and conditional generation fidelity measurements. This comprehensive evaluation approach assesses the performance of \glspl{dpm} across diverse task types, environmental conditions, and robotic configurations.

\subsubsection{Recent Advances and Integration}

Recent developments include the introduction of 3D Diffusion Policy (DP3), which enables point cloud–driven spatial generalization, and its integration with \gls{rl} through \gls{dppo} for stable fine-tuning under sparse or noisy reward conditions. The denoising-based training objective supports robust convergence even with limited expert demonstrations, ensuring scalability in data-constrained settings. Advances in diffusion-based industrial design \cite{leng2025diffusion} demonstrate that strategic noise scheduling and latent space optimization significantly enhance generation quality. For Diffusion Policy Models, adapting noise schedules to task horizons reduces inconsistencies in long action sequences, while structured latent spaces enforce constraints like workspace limits and contact forces without post-processing. In AIGC-driven manufacturing \cite{leng2026aigc}, these models generate both control policies and synthetic data for digital twins, requiring evaluation metrics that prioritize production efficiency, computational cost, and cross-scenario adaptability alongside task success.
 Diffusion-based policies have also been integrated into modern VLA architectures as action decoders. A notable example is UC Berkeley’s Octo model~\cite{team2024octo}, which leverages diffusion policies to generate continuous joint trajectories and achieve rapid adaptation. Empirical evaluations demonstrate that VLA architectures enhanced with diffusion components outperform standalone VLA models, achieving over 80\% task success in robotic manipulation, compared to fewer than 40\% for vanilla implementations. Fine-tuned VLA–diffusion hybrids consistently outperform imitation learning and direct behavioral cloning baselines, with conditioning mechanisms ensuring that generated trajectories remain both visually and proprioceptively grounded.

\subsection{World Foundation Models}

\begin{figure*}[h]
    \centering
    \includegraphics[trim=0cm 15cm 0cm 2.7cm, clip, width=\linewidth]{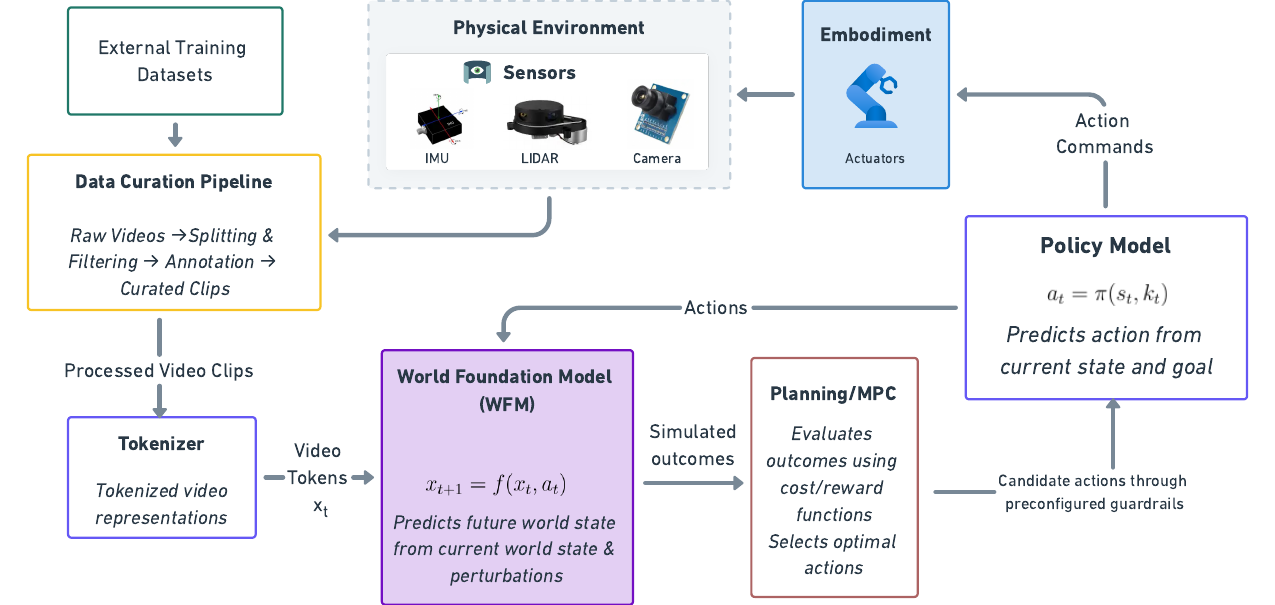}
    \caption{WFM training and closed-loop control architecture. Raw sensor streams and video data are tokenized to train the WFM to predict future states ($x_{t+1} = f(x_t, a_t)$). During control, the planner/MPC queries the WFM to generate simulated rollouts and select optimal candidate actions, which the policy model executes on the robot.}
    \label{fig:wfm}
\end{figure*}

\glspl{wfm} are learned predictive models of environment dynamics that operate on observed or latent representations rather than explicit physical states. They simulate future images, videos, or latent world representations conditioned on the current state and candidate actions, enabling counterfactual evaluation of environment evolution. By generating such simulations before execution, WFMs support planning, validation, and synthetic data generation in controlled computational settings, reducing physical risk during development and deployment when combined with appropriate constraints and verification mechanisms.

Formally, let $s_t$ denote raw observations at time $t$ (e.g., video and sensor streams) and let $x_t=\mathrm{enc}(s_t)$ denote a tokenized observation or latent world representation---not necessarily the full Markov state of the underlying physical system. Given an action $a_t$, a single-step world prediction is expressed as
\begin{equation}
x_{t+1} = f(x_t, a_t).
\end{equation}
In practice, WFMs are queried to produce \emph{rollouts}, defined as simulated sequences of future representations generated by recursively or jointly applying the model over a finite horizon. Given a candidate action sequence $a_{t:t+H-1}$, the model predicts an $H$-step rollout
\begin{equation}
x_{t+1:t+H} = f(x_t, a_{t:t+H-1}),
\end{equation}
which represents a hypothetical future trajectory of the environment under that action sequence. Depending on design and application, rollouts may correspond to predicted image frames, video sequences, latent trajectories, or sensor-level outputs (e.g., depth, segmentation, or occupancy). Because prediction errors compound over longer horizons, WFMs are typically queried for short-horizon rollouts and used with frequent replanning in receding-horizon settings.

WFMs are trained on curated datasets comprising external logs (e.g., demonstrations or driving data) and on-robot interaction data. Training begins with data curation: temporal streams from multiple sensors are synchronized, segmented into episodes or clips, and filtered for quality, with optional annotation when task-relevant labels are required. Raw sensory inputs — primarily video, but potentially including proprioceptive measurements, LiDAR scans, and inertial data — undergo tokenization to compress high-dimensional observations into compact representations. For video-centric WFMs, specialized video tokenizers compress spatiotemporal data while preserving salient semantic and dynamical content; for multimodal architectures, modality-specific encoders produce tokens that are subsequently fused into a unified representation space. The training objective typically involves next-step or next-horizon prediction conditioned on past observations and (when available) actions, enabling self-supervised learning of environment dynamics from interaction trajectories.

WFMs are commonly implemented using two generative architectural paradigms. \emph{Autoregressive} models predict future tokens sequentially in temporal order, making them well-suited for discrete latent representations and enabling fine-grained conditioning and controllability . \emph{Diffusion-based} models generate predictions through iterative denoising, representing multimodal future distributions and often achieving high-fidelity samples, albeit with higher inference cost \cite{ho2022video}. In practice, the choice between autoregressive and diffusion approaches depends on application requirements: autoregressive models favor computational efficiency and sequential controllability, while diffusion models frequently excel in generating high-quality, diverse predictions when multiple plausible futures exist \cite{ho2022video}.

As illustrated in Fig.~\ref{fig:wfm}, WFMs function within a closed-loop control architecture that integrates prediction, planning, and policy execution. A planner or \gls{mpc} system proposes candidate action sequences via sampling or optimization methods. For each candidate $a_{t:t+H-1}^{(i)}$, the WFM generates predicted rollouts $x_{t+1:t+H}^{(i)}$; multiple rollouts may be sampled per candidate to account for predictive uncertainty. The planner evaluates simulated outcomes using cost or reward functions---potentially including learned reward models---combined with constraint penalties encoding safety requirements and operational guardrails. The action sequence yielding the optimal evaluation score is selected, and its first action $a_t^*$ is executed under a receding-horizon strategy, wherein replanning occurs at subsequent timesteps to mitigate error accumulation. To amortize planning at runtime, a policy model can be trained to approximate planner-selected actions via distillation, learning a mapping $\pi:(x_t,g_t)\mapsto a_t$, where $g_t$ denotes task or goal conditioning; alternatively, model-based reinforcement learning can leverage WFM rollouts to optimize value functions or directly improve policy performance through simulated experience. Physical execution yields new observations that are logged and fed back into the curation pipeline, enabling periodic refinement through continued training on deployment data.

\subsubsection{Environmental Simulation and Data Generation}

WFMs complement traditional \glspl{dt}, which maintain virtual replicas of physical systems using structured simulation environments (often physics-based) synchronized with real-world data streams \cite{long2025survey,dihan2024digital}. While digital twins provide interpretable, constraint-driven modeling, WFMs offer learned, data-driven rollout capabilities that can capture complex dynamics and observation statistics that are difficult to model analytically. Hybrid pipelines increasingly combine both approaches: digital twins supply structural priors and hard constraints, while learned components model residual dynamics and sensor observations or generate edge-case variations \cite{ensinger2024learning,heiden2021neuralsim}. The Sim2Real gap---discrepancies between simulated and real-world behavior---arises from limitations in modeling contact dynamics, friction, deformable materials, sensor noise, lighting variation, and actuation delays \cite{muller2022self}. WFMs address aspects of this gap by learning observation-space prediction directly from real data, enabling more realistic visual and sensory outputs, though they do not eliminate fundamental uncertainty in predictive modeling.

To address the uncertainty of purely data-driven approximations and to strengthen the realism of generated environments, recent frameworks are increasingly integrating Physics-Informed Machine Learning\cite{li2025pin}. Embedding governing equations and conservation laws directly into the loss functions or latent space of WFMs ensures that generated simulations follow physics rather than just statistical approximations. This incorporation is critical in intelligent manufacturing, where accurate modeling of thermodynamics and material properties can significantly reduce sim-to-real discrepancies \cite{leng2025physics}. This alignment with physical constraints bridges the trust gap in synthetic data, enabling its safe deployment in critical systems like autonomous vehicles and industrial robots.

A key application of WFMs is controllable synthetic data generation for downstream perception and control systems. By conditioning predictions on varied initial states and action sequences, WFMs can amplify rare events-such as near-collision scenarios in autonomous driving or object failures in manipulation-transforming limited observed instances into diverse scenario families across environmental conditions (e.g., lighting, weather, and scene configurations) \cite{agarwal2025cosmos}. These synthetic datasets can be used by perception models for recognition tasks, by policy networks for behavioral cloning or offline reinforcement learning, and by safety validation frameworks for stress testing under controlled scenario distributions.

WFMs can be integrated into continual learning pipelines where real-world deployment data informs model updates. Common feedback mechanisms include: (i) distillation-based policy learning, where policies imitate actions selected by WFM-guided planners; (ii) model-based reinforcement learning, where WFM rollouts support value or policy improvement through simulated experience, with periodic grounding on real interaction data to mitigate model bias and drift; and (iii) calibration loops in hybrid WFM--digital twin systems that adjust simulator parameters based on real-world observations to improve correspondence between predicted and actual outcomes. These feedback strategies enable adaptation to changing environments and tasks, but require careful validation to prevent regressions in safety-critical settings.

\paragraph{Case study: NVIDIA Cosmos platform.}
NVIDIA Cosmos exemplifies a WFM platform designed for synthetic data generation and world simulation workflows in physical AI applications \cite{agarwal2025cosmos}. The platform comprises three primary components: Cosmos Predict generates future world states as video from multimodal prompts for scenario forecasting and planning; Cosmos Transfer converts structured inputs (e.g., LiDAR point clouds, segmentation maps, depth maps, and HD maps) into photorealistic controllable scenes; and Cosmos Reason provides a reasoning-capable vision-language model to assist with scenario analysis, curation, and annotation in physical AI contexts \cite{agarwal2025cosmos}.

{
\renewcommand{\arraystretch}{1.1} 
\begin{table*}[tb]
\caption{Comparison of different GPAI model types for embodied AI systems.}
\label{tab:model_comparison}
\centering
\small
\setlength{\tabcolsep}{4pt} 

\begin{tabularx}{\textwidth}{@{}%
  >{\centering\arraybackslash}P{1cm}         
  >{\RaggedRight\arraybackslash}p{1.8cm}     
  >{\RaggedRight\arraybackslash}p{1.9cm}
  >{\RaggedRight\arraybackslash}p{1.6cm}        
  Y                                         %
  Y
  >{\RaggedRight\arraybackslash}p{3.2cm}
  >{\RaggedRight\arraybackslash}p{2cm}
@{}}
\toprule
\rowcolor[HTML]{E0E0E0} 
{\textbf{Model Type}\par} &
{\textbf{Core Functionality}\par} &
{\textbf{Example Models}\par} &
{\textbf{Embodiment}\par} &
{\textbf{Data}\par} &
{\textbf{Learning Process}\par} &
{\textbf{Adaptability}\par} &
{\textbf{Applications}\par} \\
\midrule
RFMs & Generalization, Planning &
PaLM-E~\cite{driess2023palm}, Gato~\cite{reed2022generalist} &
All robot types &
Large-scale multi-robot datasets (e.g., Open X-Embodiment, DROID) &
Self-supervised, few-shot learning &
 Zero-shot/few-shot transfer across tasks and robot morphologies &
Robotics generalization, planning, and control \\
\addlinespace
VLAs & Multimodal Perception/Action &
RT-1~\cite{brohan2022rt}, RT-2~\cite{zitkovich2023rt}, OpenVLA~\cite{kim2024openvla} &
Manipulators, mobile &
Multimodal sensory data (vision, language, proprioception) &
Multimodal learning, supervised fine-tuning &
Real-time policy adjustment by grounding abstract instructions and feedback streams &
Robot perception and multimodal interaction \\
\addlinespace
LBMs & Human-like Behavior &
Meta Motivo~\cite{tirinzoni2025zero} &
Humanoids, collaborative &
Human interaction and behavior datasets &
Imitation learning, RL &
Generalizes social interaction policies and transfers learned social and behavioral cues &
Human-robot collaboration and social robotics \\
\addlinespace
Diffusion Policy & Generative Action Control &
Diffusion Policy~\cite{chi2023diffusion} &
Manipulators &
Manipulation task data, expert demonstrations &
Diffusion models \& RL &
Focuses on error recovery and handling uncertainty in the action space &
Robotic manipulation and control \\
\addlinespace
WFMs & World Simulation/Prediction &
NVIDIA Cosmos~\cite{agarwal2025cosmos} &
Simulation (\gls{dt}) &
Simulated world data, \glspl{dt} &
Simulation-based training, world model learning, physics-informed learning &
Enhances sim-to-real transfer and adapts to novel environmental dynamics &
Training and evaluation of embodied AI systems \\
\bottomrule
\end{tabularx}
\end{table*}
}

\subsubsection{Evaluation and Benchmarking}

\paragraph{Evaluation objectives.}
WFM evaluation must assess multiple dimensions of capability: generation quality and temporal stability, controllability under conditioning signals (e.g., actions or textual prompts), physical plausibility (e.g., object permanence and basic dynamical consistency), and downstream utility for training and decision-making. Benchmarks therefore combine perceptual metrics, physics-oriented failure detection, task-oriented performance measures, and human preference assessments to characterize strengths and limitations across these axes.

\paragraph{WorldScore benchmark.}
WorldScore \cite{duan2025worldscore} provides a structured framework encompassing controllability, quality, and dynamics across 3,000 next-scene prediction scenarios spanning static and dynamic scenes in both indoor and outdoor settings. Controllability is quantified via camera controllability, object controllability, and content alignment, enabling unified evaluation of diverse approaches under explicit layout and prompt specifications.

\paragraph{WorldModelBench.}
WorldModelBench \cite{li2025worldmodelbench} evaluates video generation models as world models on application-driven domains, explicitly measuring instruction-following and physics-adherence dimensions beyond generic video quality. It includes 350 image-and-text condition pairs spanning multiple domains and is supported by 67K crowd-sourced human labels to assess 14 frontier models, highlighting subtle violations such as physically implausible size changes that breach mass conservation.

\paragraph{WorldSimBench.}
WorldSimBench \cite{qin2024worldsimbench} introduces a two-part protocol comprising Explicit Perceptual Evaluation and Implicit Manipulative Evaluation. The explicit component uses the HF-Embodied dataset (over 35,000 annotated tuples with multi-dimensional human feedback) to evaluate perceptual quality and condition consistency. The implicit component evaluates downstream utility by measuring whether generated situation-aware videos can support embodied action through video-to-action conversion, reporting action-level performance across embodied scenarios including open-ended environments, autonomous driving, and robot manipulation.

\subsection{Taxonomy Synthesis and Discussion}

The comparative analysis in Table~\ref{tab:model_comparison} illustrates how the GPAI model classes correspond to stages in the robotic control pipeline. \glspl{rfm} and \glspl{vla} models span perception, planning, and action, supporting broad generalization. In contrast, \glspl{wfm}, \glspl{lbm}, and \glspl{dpm} address specialized functions: WFMs emphasize predictive simulation and data generation, LBMs enable nuanced human-like interaction, and Diffusion Policies provide robust low-level control for manipulation. This mapping underscores the tradeoff between generality and specialization, linking model design to embodiment, data demands, learning strategies, adaptability, and downstream applications.

The taxonomy organizes the generative physical AI landscape by assigning core model functions to stages of the control pipeline. It highlights a shift from modular, task-specific systems toward integrated, end-to-end architectures. RFMs and VLAs exemplify this integration, spanning sensing to action by training on large, diverse datasets. Specialized models remain vital: WFMs excel in world modeling and synthetic data generation, LBMs generate fine-grained behaviors, and Diffusion Policies deliver reliable low-level control. A key trend is the increasing reliance on data-centric learning, where training data, robot interaction logs, multimodal web corpora, or curated demonstrations define a system's capability. While categories remain distinct, convergence is emerging: specialized strengths of Diffusion Policies and LBMs are expected to integrate within generalist RFMs and VLAs. This trajectory points to hybrid systems that combine broad reasoning with specialized control, enabling robots with greater adaptability, competence, and autonomy.

\section{Applications of GPAI}
\label{sec:applications}

\begin{figure}[t]
    \centering
    \includegraphics[trim=0cm 13cm 0cm 4.6cm, clip, width=\columnwidth]{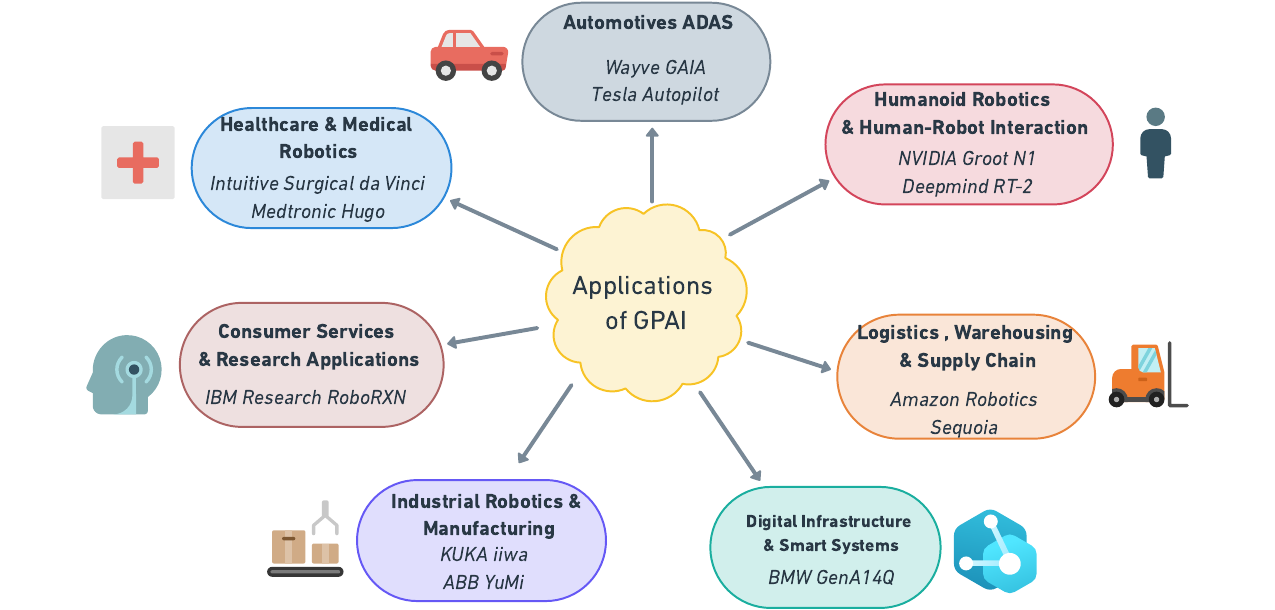}
    \caption{Applications of \gls{gpai} across domains including ADAS, robotics, healthcare, humanoids, assistive technologies, manufacturing, and digital twins.}
    \label{fig:applications}
\end{figure}

GPAI is driving transformative innovation across a wide range of industries by unifying perception, reasoning, and physical interaction.Figure~\ref{fig:applications} summarizes key GPAI domains, including autonomous driving, robotics, healthcare, humanoids, manufacturing, and digital twins. These applications illustrate GPAI’s role in linking digital intelligence with physical action. The following subsections present representative real-world examples.

\subsection{Autonomous Vehicles and Advanced Driver Assistance Systems}

In the automotive sector, GPAI underpins both \gls{adas} and fully autonomous vehicles by integrating perception, prediction, and planning into cohesive decision-making pipelines. These systems must interpret complex traffic environments, anticipate the behaviors of vehicles and pedestrians, and generate safe and efficient trajectories in real-time. A key enabler of this progress is the emergence of WFMs, which generate diverse and realistic driving scenarios to accelerate training and validation while reducing dependence on costly real-world data acquisition~\cite{agarwal2025cosmos}.

Recent advances by Wayve exemplify this paradigm through the GAIA family of generative world models, which act as neural simulators for autonomous driving. GAIA-1~\cite{hu2023gaia} formulates world modeling as an autoregressive sequence prediction task over tokenized video, text, and action inputs, combining large-scale transformers with video diffusion decoders to produce realistic, controllable driving rollouts. This architecture enables the synthesis of rare but safety-critical events that are impractical to capture at scale in naturalistic datasets. Its successor, GAIA-2~\cite{russell2025gaia}, extends this approach through a latent diffusion framework capable of generating temporally consistent, multi-view videos across geographically diverse domains. Structured conditioning on ego-vehicle dynamics, agent interactions, and road semantics enables GAIA-2 to generate scenarios that range from trajectory forecasting to fully synthetic environments, including out-of-distribution conditions. Collectively, these models demonstrate how generative simulation provides a scalable and systematic pathway for robust policy validation in autonomous vehicles.

Complementary to generative simulation, Tesla’s Autopilot adopts an end-to-end learning methodology~\cite{lan2023end}, wherein a unified neural network is trained directly on large-scale fleet data. This fleet learning paradigm establishes a powerful data flywheel, continuously collecting novel driving situations, enabling frequent model retraining, and propagating improvements across the global fleet via over-the-air (OTA) updates. Such a vision-centric approach leverages the scale of real-world human driving data to enhance generalization and adaptability. Supporting these data- and compute-intensive paradigms, the NVIDIA DRIVE platform~\cite{nvidia2025drive} delivers the high-performance hardware and software stack necessary for real-time deployment. By fusing specialized processors with a comprehensive \gls{sdk}, DRIVE enables the computational throughput required for multi-sensor fusion, perception, and motion planning under stringent latency constraints, thereby making the deployment of GPAI-driven models feasible in production vehicles.

\subsection{Industrial Robotics and Manufacturing}
Robotic manipulators powered by GPAI perform assembly, packaging, and inspection tasks with unprecedented dexterity and adaptability in industrial automation settings~\cite{liu2022robot}. Foundation models enable zero-shot generalization, allowing robots to handle novel objects and tasks without retraining, thereby reducing downtime and engineering effort\cite{sai2024generative}. Diffusion policy models generate temporally coherent action sequences, thereby improving performance in long-horizon manipulation tasks that require sustained coordination and precision\cite{yang2025automated}. ABB Robotics has deployed over 500,000 industrial robots globally~\cite{abb_yumi_dualarm}, with YuMi collaborative robots incorporating vision and force sensing capabilities. KUKA's IIWA robots utilize impedance control and machine learning for adaptive manufacturing processes~\cite{serrano2023scalable}. RNB Cosméticos employs six Universal Robots UR10 cobots with torque sensors for end-of-line palletizing of cosmetic products, handling 7kg packages at six per minute across 350+ product variants without safety barriers, demonstrating collaborative human-robot operation in constrained spaces while improving ergonomics and production efficiency.

\subsection{Healthcare and Medical Robotics}
Healthcare applications showcase GPAI's transformative potential through surgical robots that interpret multimodal data including video, force, and language inputs to plan procedures and adapt to intraoperative changes with autonomous assistance capabilities. Multimodal foundation models enhance robots' ability to handle unpredictable living tissue dynamics, thereby reducing the surgeon's burden while maintaining safety and precision~\cite{khan2025surgical}. These systems learn from expert demonstrations and synthetic data, accelerating skill acquisition and improving patient outcomes through continuous learning and adaptation~\cite{wang2024surgical}.

Intuitive Surgical's da Vinci system has enabled surgeons to conduct over 14 million minimally invasive procedures with enhanced precision, dexterity, and control across diverse medical specialties~\cite{celotto2024vinci}. Clinical outcomes show significant advantages including lower conversion rates, fewer surgical site infections, and reduced pain compared to traditional surgery, with a remarkably low 0.12\% mortality rate across over 10,000 procedures~\cite{koh2018efficacy}. Medtronic's Hugo platform offers modular robotic-assisted surgery systems that support surgeons through detailed procedure planning, real-time 3D imaging, and instrument guidance for complex operations~\cite{prata2023state}. The system has achieved exceptional clinical results with a 98.5\% surgical success rate in FDA trials, significantly exceeding the 85\% benchmark. The platform's modular design enables faster surgeries, improved ergonomics for surgeons, and reduced invasiveness, leading to shorter hospital stays and lower healthcare costs. In assistive technologies, Cyberdyne's HAL (Hybrid Assistive Limb) suits utilize bioelectric signals, with over 10,000 treatment sessions completed globally, demonstrating GPAI's capacity for personalized assistance through exoskeletons and prosthetics that adapt to individual users and their environments. Clinical trials confirm HAL's efficacy with patients walking 10\% farther in 2-minute walk tests after treatment, showing significant improvements in walking and balance that last for weeks~\cite{nakajima2021cybernic}. The 15kg exoskeleton system provides measurable therapeutic benefits for patients with neuromuscular diseases, stroke recovery, and mobility impairments through nine 40-minute treatment sessions.

\subsection{Humanoid Robotics and Human-Robot Interaction}
Humanoid robots integrate \glspl{llm}, \glspl{vlm} and \gls{rl} to understand language instructions, plan complex tasks, and interact safely with humans and environments~\cite{sheng2025comprehensive}. These systems utilize robot state representations as observation tokens encoding sensor values, gripper positions, and joint angles, while employing diffusion models to predict multimodal action sequences that handle temporal consistency through receding horizon control.

NVIDIA's Groot N1 model processes multimodal inputs including vision, language, and proprioceptive robot state through a \gls{vla} transformer backbone, employing a diffusion policy head that progressively refines predicted action sequences to produce smooth and reliable robot movements~\cite{bjorck2025gr00t}. This 2.2B parameter model achieves an inference time of 64ms, with \gls{vlm} operating at 10Hz and the Diffusion Transformer generating actions at 120Hz. This dual system architecture enables robots to reason about environments while generating fluid and real-time motor actions, significantly enhancing the capabilities of humanoid robots. Boston Dynamics' Atlas demonstrates advanced locomotion through \gls{rl}, with a focus on developing sim-to-real mobility policies and whole-body locomotion and manipulation capabilities. Its model trains on 150 million simulations per maneuver to achieve human-like agility~\cite{kuindersma2016optimization}. This approach enables zero-shot transfer from simulation to reality, allowing Atlas to perform complex locomotion tasks without extensive real-world training data. Agility Robotics' Digit has been deployed in Amazon warehouses for package handling tasks~\cite{Dresser2023AmazonRobotics}. The robot handles repetitive and heavy tasks such as bulk material handling and tote recycling. Tesla’s Optimus extends this landscape by demonstrating autonomous household assistance, performing tasks such as ironing, cleaning, and cooking at nearly twice the operational speed of teleoperated systems~\cite{kalil2025tesla}. Its training leverages large-scale online video corpora for imitation learning, enabling rapid skill acquisition and reducing the need for extensive real-world demonstrations.

Human-Robot Interaction (HRI) applications leverage LLMs and VLMs integration to enable robots to understand and respond to natural language, gestures, and emotional cues for intuitive communication. Google DeepMind’s RT-2 exemplifies this trend by extending vision-language understanding to robotic control. It achieves a 62\% success rate on unseen tasks compared to RT-1’s 32\%, highlighting significant gains in generalization and semantic reasoning~\cite{brohan2022rt, zitkovich2023rt}. By transferring knowledge from large-scale web data, RT-2 allows robots to recognize and act in novel situations without explicit task-specific training, representing a step forward in robotic adaptability. Toyota Research Institute pursues a complementary direction through fleet learning, where experience gained by one robot is shared across a distributed cohort. This approach enables robots to adapt to individual user preferences while accelerating collective learning. Recent work explores diffusion-based training methods to improve sample efficiency, with applications ranging from home assistance in kitchen environments to industrial and factory robotics.

\subsection{Logistics, Warehousing, and Supply Chain Management}
In logistics and warehouse automation, GPAI-powered robotic systems enable dynamic task allocation for picking, packing, and transportation, adapting in real-time to fluctuations in inventory levels and demand patterns~\cite{dehghan2023dynamicagvtaskallocation}. Foundation models enhance these systems by allowing robots to manipulate diverse objects and handle unstructured scenarios with minimal supervision, increasing robustness where conventional automation often fails. Amazon Robotics exemplifies the scale of this transformation, having deployed more than 750,000 robots across its global operations network since 2012. Recent initiatives, such as the Sequoia inventory management platform, utilize AI-driven path planning and task scheduling to enhance fulfillment center productivity by approximately 25\%~\cite {greenawalt_robots_2025}. Similarly, Walmart employs proprietary GPAI-based systems to optimize supply chain efficiency. Its route optimization software reduces redundant travel and improves trailer packing configurations, preventing an estimated 30 million unnecessary driving miles and cutting carbon emissions by 94 million pounds annually~\cite{walmart_route_optimization_2024}. Beyond routing, Walmart is integrating GPAI into inventory management and deploying autonomous robotic forklifts throughout its distribution centers, significantly accelerating product handling and warehouse throughput.  

\subsection{Digital Infrastructure and Smart Systems}
Digital twins and smart spaces extend GPAI capabilities by generating high-fidelity virtual replicas of factories, warehouses, and urban environments. These replicas enable predictive maintenance, process optimization, and scenario-based planning, while also serving as synthetic data sources that accelerate the training and evaluation of GPAI models. Such approaches reduce both the financial and operational risks associated with real-world deployment~\cite{agarwal2025cosmos}. A prominent industrial example is the BMW Group’s Virtual Factory initiative, which has been scaled across digital twins of more than 30 global production sites and is projected to reduce production planning costs by up to 30\%~\cite{bmw2025virtualfactory}. This virtual-to-physical integration is exemplified at the company’s Regensburg plant, where AI-enabled quality control systems monitor approximately 1,400 vehicles produced daily. The facility manufactures a new car every 57 seconds and incorporates the “GenAI4Q” pilot system to generate tailored inspection recommendations for each vehicle~\cite{Graser2025GenAI4Q}. Collectively, these implementations demonstrate how digital twins, combined with GPAI, can streamline pre-production planning while simultaneously enhancing active manufacturing operations, thereby establishing an integrated smart manufacturing ecosystem that spans simulation, validation, and execution.

\subsection{Consumer Services and Research Applications}
Consumer and service robotics increasingly integrate GPAI to support home automation, where robots perform household tasks, assist with daily living, and provide companionship. In retail and hospitality, service robots employ GPAI for natural interaction, autonomous navigation, and context-aware task execution, thereby enhancing customer engagement and operational efficiency~\cite{lisondra2025embodied}. In research domains such as material science, chemistry, and product design, GPAI’s generative capabilities accelerate discovery by enabling the creation of novel material structures, prototypes, and chemical compounds~\cite{m2024augmenting, das2024overview}. A representative example is IBM Research’s RoboRXN platform, which combines AI-driven simulation with robotic synthesis to reduce compound development cycles from weeks to hours. By guiding experimental design and minimizing reliance on trial-and-error methods, RoboRXN demonstrates how GPAI can substantially increase the efficiency and scalability of laboratory research~\cite{o2021ai}.

\section{Challenges}
\label{sec:challenges}

\subsection{Data Scarcity and Requirements}
A primary challenge in developing GPAI systems is the immense data requirement for training robust world models and foundation models in robotics. Unlike text- or image-based generative AI, GPAI models demand datasets that are not only vast and diverse but also physically grounded, encoding cause–and–effect relationships, multi-step interactions, and the fine-grained dynamics of manipulation. Data scarcity becomes particularly acute in high-stakes, safety-critical edge cases, which are both rare in natural settings and costly to reproduce in controlled experiments. The collection, curation, and annotation of such datasets are resource-intensive, often involving specialized sensors, complex hardware testbeds, and extensive human oversight~\cite{hakami2024strategies}. These barriers make large-scale, high-quality physical datasets a persistent bottleneck for advancing GPAI. However, emerging approaches such as self-supervised learning, synthetic data generation, and transfer learning from simulation offer pathways to reduce dependence on large-scale physical datasets~\cite{jeong2020selfsupervised}.In particular, frameworks integrated with digital twins provide a structured mechanism for synthetic data generation within closed-loop simulation environments\cite{chen2024aigc}.

\subsection{Bias and Ethical Concerns}
Data challenges in GPAI extend beyond scarcity to the pervasive issue of systemic bias. Limited diversity or skewed representation in training datasets can cause models to exhibit brittle, prejudiced, or unsafe behaviors when deployed in the real world\cite{kumar2025peeping}. For embodied systems, such biases may translate into discriminatory or hazardous physical actions, representing not only a technical flaw but also a profound ethical failure~\cite{hanna2025ethical}. The implications reach further: biased AI creates legal and regulatory liabilities, as the deployment of unfair systems can conflict with anti-discrimination laws and expose organizations to financial or reputational damage. To mitigate these risks, emerging governance frameworks, such as the European Union’s AI Act~\cite{smuha2025regulation}, mandate transparency, bias detection, and mitigation strategies for high-risk AI. Addressing bias is therefore simultaneously an ethical responsibility, a regulatory requirement, and a precondition for trustworthy GPAI.
Beyond data-level bias mitigation, the deployment of GPAI systems should align with emerging Industry 5.0 principles, which emphasize human-centricity, sustainability, and resilience\cite{rovzanec2023human}. GPAI should augment rather than replace human capabilities, particularly in collaborative domains such as smart manufacturing and surgical robotics. To operationalize these principles, evaluation frameworks must incorporate socio-technical metrics, including explainability, human trust, safety compliance, and inclusivity across diverse user populations. Such holistic evaluation helps ensure that autonomous systems remain ethically aligned with societal values, especially in safety-critical deployments.

\begin{table*}[tb]
\caption{Summary of key limitations, challenges, risks, and emerging solutions in GPAI systems.}
\label{tab:challenges}
\renewcommand{\arraystretch}{1.3}
\setlength{\tabcolsep}{6pt}
\centering
\small
\begin{tabularx}{0.99\linewidth}{
    >{\raggedright\arraybackslash}X
    >{\raggedright\arraybackslash}p{5 cm}
    >{\raggedright\arraybackslash}p{3.2cm}
    >{\raggedright\arraybackslash}p{2.2cm}
    >{\raggedright\arraybackslash}p{2.8cm}
}
\toprule
\rowcolor{headergray}
\textbf{Limitation} & 
\textbf{Description} & 
\textbf{Key Challenges} & 
\textbf{Safety/Risk Implications} & 
\textbf{Proposed Solutions} \\
\midrule

\textbf{Data Scarcity \& Requirements} & 
Requires vast, physically-grounded data that is resource-intensive to collect. & 
Complex hardware setups, human oversight, rare edge cases & 
Data gaps in safety-critical contexts & 
Self-supervised, few-shot, transfer learning \\

\textbf{Bias \& Ethical Concerns} & 
Biased data can lead to unsafe and discriminatory actions, as well as regulatory risks. & 
Legal risks, penalties, underrepresented groups & 
Unsafe behavior, ethical failure & 
Governance, bias detection, mitigation \\
\textbf{Hardware \& Real-Time Constraints} & 
Computational intensity conflicts with real-time processing on edge devices. & 
Latency, low power or memory, optimization tradeoffs & 
Delays or errors in split-second decisions & 
Model compression, edge AI, dedicated hardware \\

\textbf{Simulation-to-Real Transfer} & 
The "reality gap" between clean simulations and the complex real world. & 
Dynamic variations, friction, noise & 
Unsafe transfers to physical systems & 
Digital twins, robust and differentiable simulators \\

\textbf{Generalization \& Reasoning} & 
Poor contextual reasoning impairs understanding of physical actions. & 
Misinterprets relations, fails on logical outcomes & 
Unsafe or flawed actions & 
Advanced reasoning models, multimodal integration \\

\textbf{Scalability \& Complexity} & 
Integrating perception, reasoning, and control modules is complex and hinders scaling. & 
Complexity rises, reliability falls & 
Hard to scale from lab to real world & 
Modular and adaptable model architectures \\


\textbf{Safety \& Robustness} & 
Must achieve near-perfect reliability, as failures can cause real-world harm. & 
Critical failure modes not seen in digital systems & 
Harm to humans and property & 
Rigorous safety checks, monitoring, fail-safes \\

\textbf{Transparency \& Explainability} & 
"Black box" decision processes make auditing and error diagnosis difficult. & 
Hard to explain or hold accountable & 
Barrier for critical systems & 
Explainable AI, interpretable models \\

\textbf{Energy Efficiency} & 
Training demands massive computational power, which has a significant carbon footprint. & 
Sustainability challenges & 
Environmental impact & 
Efficient algorithms, green computing \\


\bottomrule
\end{tabularx}
\end{table*}

\subsection{Hardware and Real-Time Constraints}
GPAI systems are limited by hardware and computational constraints that directly impact real-world deployment. Generative models are often computationally intensive, making it challenging to satisfy the strict latency requirements of real-time perception and control. This becomes especially critical in robotics, where even minor delays can compromise both performance and safety. Physical systems must operate under strict computation and communication constraints  imposed by the edge environment\cite{rodrigues2019machine}.This constraint is particularly acute for VLAs, where large vision-language backbones require significant memory and processing power, limiting their deployment as fast reactive controllers on embedded hardware.Diffusion Policy Models face similar latency challenges. Their iterative denoising process requires multiple forward passes per control decision, which can conflict with the sub-50ms response times needed in fast manipulation tasks. Deploying GPAI on edge devices introduces further constraints: limited compute power, memory, and energy budgets restrict the feasible model size and complexity\cite{sai2024device}. These bottlenecks typically necessitate aggressive compression or optimization, which can reduce accuracy, degrade generalizability, and ultimately limit the reliability of embodied systems. However, emerging techniques such as model quantization, pruning, knowledge distillation~\cite{li2023model}, and hybrid edge-cloud architectures~\cite{jouini2024survey} offer promising pathways to mitigate these constraints while maintaining model performance.

\subsection{Simulation-to-Real Transfer and Uncertainty}
Bridging the “reality gap” remains one of the most persistent technical and safety hurdles in GPAI. This challenge often affects Large Behavior Models, which demonstrate impressive capabilities in simulation but have limited real-world validation beyond basic locomotion tasks, as its performance often degrades when confronted with unmodeled contact scenarios and hardware imperfections. Differentiable simulation and physics-informed training enhance fidelity, but simulations inevitably simplify the complex richness of the physical world. Subtle variations in material properties (e.g., friction, mass), nonlinear dynamics (e.g., contact physics, actuator imprecision), and pervasive sensor noise are often not fully captured. Consequently, models trained solely in simulation often overfit to these clean environments, producing behaviors that collapse under real-world uncertainty. Unlike errors in text or image generation, which are confined to digital artifacts, failures in GPAI can manifest as unsafe or damaging actions in embodied systems. Addressing this gap is therefore not merely a performance issue but a foundational requirement for building reliable and trustworthy physical AI. Techniques such as domain randomization, sim-to-real transfer learning, and iterative reality gap refinement are actively narrowing this divide~\cite{jeong2020selfsupervised}.

\subsection{Generalization, Prompt Sensitivity, and Reasoning Challenges}
Current GPAI systems face persistent challenges in higher-level contextual reasoning, temporal prediction, and robust generalization.For Robot Foundation Models specifically, cross-platform deployment introduces embodiment mismatch with differences in actuator dynamics, joint limits, and sensing configurations. This can cause policies to behave unpredictably on new robots . Continual fine-tuning on new domains can also lead to catastrophic forgetting of previously learned skills. Failures often arise when robots misinterpret object relationships or overlook the long-term consequences of their actions. Similar to hallucinations in large language models, such reasoning errors can cascade into unsafe behaviors when embodied in the physical world. Prompt sensitivity further undermines reliability: even slight variations in user input can yield inconsistent or hazardous outputs~\cite{razavi2025benchmarkingpromptsensitivitylarge}. Beyond task-specific generalization, the field continues to grapple with the more complex problem of open-ended generalization~\cite{hughes2024openendednessessentialartificialsuperhuman}, i.e., the ability of models to adapt continually in dynamic environments, infer novel object affordances, and acquire new skills without explicit retraining. Achieving this capacity is central to building GPAI systems that are not only accurate in controlled settings but also resilient, safe, and versatile in the open world. Advances in multi-task learning, meta-learning, and continual adaptation frameworks show promise in addressing these generalization bottlenecks.

\subsection{Scalability and System Complexity}
Scaling GPAI systems extends beyond improving individual models to addressing the integration of heterogeneous subsystems, including perception, reasoning, planning, and control. Ensuring that these modules operate reliably and coherently is a nontrivial challenge, as failures in one layer can propagate and destabilize the entire system. With increasing task complexity and environmental variability, the reliability burden grows exponentially~\cite{wan2025generative}. Integration complexity, therefore, remains a central barrier to scaling GPAI from controlled laboratory prototypes to robust deployment in open and unstructured real-world settings. Modular architectures, hierarchical planning systems, and improved integration frameworks offer pathways to manage this complexity at scale~\cite{zhang2025review}.


\subsection{Robustness, Safety, and Failure Modes}
GPAI systems extend AI decision-making into the physical world, where failures can cause direct harm to people, infrastructure, and the environment. Unlike software-only domains, deployment in physical settings requires reliability thresholds that approach near-perfect performance. Errors in reasoning, breakdowns in system integration, or insufficiently fine-grained control can propagate through tightly coupled components, leading to cascading failure modes that are difficult to predict and costly to contain. Addressing these risks requires rigorous safety evaluation, continuous monitoring under real-world conditions, and the design of robust, fail-safe mechanisms across the model lifecycle, from simulation and training to field deployment~\cite{xing2025towards}.

At the control level, model design choices can further constrain safety and performance. Precise manipulation tasks demand fine motor control, yet many vision-language-action models discretize actions into fixed token vocabularies (e.g., 256 bins per dimension) to improve training manageability. While effective for learning, this reduced control resolution can cause small motion discontinuities that hinder smooth execution of finely calibrated operations. For diffusion-based controllers, an additional concern is the lack of formal stability and safety guarantees. These stochastic policies are typically validated empirically, and there are no widely adopted methods to certify their behavior under disturbances in safety-critical settings.

Runtime safety enforcement offers a partial mitigation. Techniques such as Control Barrier Function-based filters can impose state-space constraints on learned policies at each actuation step, providing formal guarantees that the system remains within a defined safe operating region even when the policy behaves unexpectedly\cite{taylor2020learning}. However, composing such filters with the stochastic, high-dimensional outputs of foundation models remains an open challenge. Existing evaluation and certification frameworks continue to emphasize task success in controlled environments and fail to adequately capture operational safety requirements, including contact force limits, failure recovery time, and energy-constrained execution~\cite{tang2024defining}. As a result, a significant gap persists between the safety assurances required for deployment in safety-critical domains and what current standards can formally guarantee.

\subsection{Transparency and Explainability}
Most GPAI models function as “black boxes,” with internal mechanisms that remain opaque to both developers and end users. This lack of interpretability complicates the diagnosis of errors, limits accountability, and erodes user confidence. The challenge is amplified in safety-critical domains, where decisions must be auditable to meet regulatory, ethical, and trust requirements. Consequently, advancing methods for interpretable and transparent GPAI is not merely a technical aspiration but a prerequisite for reliable deployment and societal acceptance~\cite{schneider2024explainable}.

\subsection{Energy Efficiency and Environmental Impact}
Training large-scale GPAI models requires substantial computational resources, resulting in high energy consumption and notable carbon emissions~\cite{jegham2025hungry}. As model sizes and training workloads continue to scale, the environmental footprint becomes a critical concern. Beyond immediate energy costs, the long-term sustainability of GPAI development raises pressing questions about balancing performance gains with ecological responsibility and carbon efficiency.


Table~\ref{tab:challenges} presents an overview of the primary challenges identified in the preceding discussion. It systematically categorizes the challenges inherent in GPAI development across critical dimensions, namely data, computation, safety, and control, while mapping each challenge to its associated risks and corresponding emerging mitigation strategies. This structured summary complements the qualitative analysis, providing a coherent reference point to inform future research directions and the development of more robust GPAI systems.

\section{Future Directions}
\label{sec:future_directions}

Addressing the current constraints of \gls{gpai} requires systematic advances in research across multiple domains. Critical research directions include developing data-efficient learning paradigms to reduce dependence on large-scale annotated datasets, designing foundation models that enable cross-embodiment generalization, and achieving computational efficiency improvements for real-time deployment on edge devices. Safety validation protocols and ethical governance frameworks enforcement are also very essential for ensuring reliable deployment. These efforts are fundamental to realizing autonomous systems with improved capability, robustness, and adaptability for practical applications.

\subsection{Data Efficiency and Learning Approaches}

Future models must learn effectively from smaller, less curated datasets. Self-supervised learning, transfer learning, and few-shot learning approaches will reduce dependence on massive labeled datasets, enabling more agile scaling to new domains and tasks. Enhanced simulation environments and digital twins will provide diverse, realistic training scenarios, helping bridge the persistent sim-to-real gap that currently limits policy transferability~\cite{Mucci2024}.

\subsection{Foundation Models and Modular Architectures}

The development of more general and modular foundation models for robotics represents a paradigm shift from task-specific architectures. Future \glspl{rfm} and \glspl{vla} will support plug-and-play adaptation across different physical embodiments and environments. Advances in reasoning architectures and multimodal integration will be crucial for addressing current reasoning failures and enabling more robust causal understanding~\cite{hu2023toward}.

\subsection{Computational Efficiency and Hardware Advances}

Simultaneously, computational efficiency remains critical for real-time inference on resource-constrained hardware. Advances in edge AI, model compression, and specialized hardware will enable the deployment of AI on mobile robots and embedded systems. Progress in fine-grained control will require breakthroughs in haptic feedback systems, advanced actuator technologies, and learned dexterity models that can match human-level precision.

\subsection{Safety Protocols}

As these systems become more autonomous, robust safety protocols and external auditing tools become essential for monitoring and validating model behavior. Comprehensive safety frameworks must specifically address reasoning failures and control precision limitations to ensure reliable operation in critical applications. Continual and interactive learning capabilities will enable systems to improve through real-world feedback and human interaction, proving particularly valuable for long-term deployment in dynamic environments. The widespread adoption of GPAI technologies necessitates clear ethical guidelines and regulatory frameworks.

\section{Conclusion}
\label{sec:conclusion}

GPAI is ushering in a new era of embodied intelligence by uniting large-scale generative models with diverse robotic platforms. This integration enables systems that can perceive, reason, and act within complex real-world environments exhibiting adaptive, closed-loop behaviors that generalize across tasks and conditions. By leveraging multimodal sensing and real-time feedback, GPAI systems continually refine their control policies, achieving greater robustness and flexibility.

Recent advances in end-to-end training and \gls{rl} have allowed GPAI models to map raw sensory data directly to executable actions. Hierarchical architectures that link high-level planning with low-level control enhance sample efficiency and scalability, supporting operations in dynamic and high-dimensional settings. Embedding physical realism into generative models remains a key frontier. Differentiable simulations and physics-aware training ensure that predicted behaviors are not only plausible but also physically consistent, a requirement for safe real-world deployment in robotics and autonomous systems. The rise of digital twins and high-fidelity simulators has expanded the available training data; however, the persistent simulation-to-real gap continues to demand robust domain adaptation and continuous real-world validation.


\bibliographystyle{IEEEtran}
\bibliography{references}

@article{li2025comprehensive,
  title={A Comprehensive Survey on World Models for Embodied AI},
  author={Li, Xinqing and He, Xin and Zhang, Le and Liu, Yun},
  journal={arXiv preprint arXiv:2510.16732},
  year={2025}
}

@article{firoozi2025foundation,
  title={Foundation models in robotics: Applications, challenges, and the future},
  author={Firoozi, Roya and Tucker, Johnathan and Tian, Stephen and Majumdar, Anirudha and Sun, Jiankai and Liu, Weiyu and Zhu, Yuke and Song, Shuran and Kapoor, Ashish and Hausman, Karol and others},
  journal={The International Journal of Robotics Research},
  volume={44},
  number={5},
  pages={701--739},
  year={2025},
  publisher={SAGE Publications Sage UK: London, England}
}

@article{kim2024openvla,
  title={Openvla: An open-source vision-language-action model},
  author={Kim, Moo Jin and Pertsch, Karl and Karamcheti, Siddharth and Xiao, Ted and Balakrishna, Ashwin and Nair, Suraj and Rafailov, Rafael and Foster, Ethan and Lam, Grace and Sanketi, Pannag and others},
  journal={arXiv preprint arXiv:2406.09246},
  year={2024}
}

@inproceedings{zitkovich2023rt,
  title={Rt-2: Vision-language-action models transfer web knowledge to robotic control},
  author={Zitkovich, Brianna and Yu, Tianhe and Xu, Sichun and Xu, Peng and Xiao, Ted and Xia, Fei and Wu, Jialin and Wohlhart, Paul and Welker, Stefan and Wahid, Ayzaan and others},
  booktitle={Conference on Robot Learning},
  pages={2165--2183},
  year={2023},
  organization={PMLR}
}

@article{agarwal2025cosmos,
  title={Cosmos world foundation model platform for physical ai},
  author={Agarwal, Niket and Ali, Arslan and Bala, Maciej and Balaji, Yogesh and Barker, Erik and Cai, Tiffany and Chattopadhyay, Prithvijit and Chen, Yongxin and Cui, Yin and Ding, Yifan and others},
  journal={arXiv preprint arXiv:2501.03575},
  year={2025}
}

@article{gat1998three,
  title={On three-layer architectures},
  author={Gat, Erann and Bonnasso, R Peter and Murphy, Robin and others},
  journal={Artificial intelligence and mobile robots},
  volume={195},
  pages={210},
  year={1998}
}

@article{sapkota2025vision,
  title={Vision-language-action models: Concepts, progress, applications and challenges},
  author={Sapkota, Ranjan and Cao, Yang and Roumeliotis, Konstantinos I and Karkee, Manoj},
  journal={arXiv preprint arXiv:2505.04769},
  year={2025}
}

@article{brohan2022rt,
  title={Rt-1: Robotics transformer for real-world control at scale},
  author={Brohan, Anthony and Brown, Noah and Carbajal, Justice and Chebotar, Yevgen and Dabis, Joseph and Finn, Chelsea and Gopalakrishnan, Keerthana and Hausman, Karol and Herzog, Alex and Hsu, Jasmine and others},
  journal={arXiv preprint arXiv:2212.06817},
  year={2022}
}

@article{liu2022robot,
  title={Robot learning towards smart robotic manufacturing: A review},
  author={Liu, Zhihao and Liu, Quan and Xu, Wenjun and Wang, Lihui and Zhou, Zude},
  journal={Robotics and Computer-Integrated Manufacturing},
  volume={77},
  pages={102360},
  year={2022},
  publisher={Elsevier}
}

@misc{abb_yumi_dualarm,
  title        = {{Dual-arm YuMi® – IRB 14000 (Collaborative Robot, ABB)}},
  author       = {{ABB}},
  howpublished = {\url{https://new.abb.com/products/robotics/robots/collaborative-robots/yumi/dual-arm}},
  year         = {2025},
  note         = {Accessed: 2025-08-25},
}

@inproceedings{serrano2023scalable,
  title={A scalable and unified multi-control framework for KUKA LBR iiwa collaborative robots},
  author={Serrano-Mu{\~n}oz, Antonio and Elguea-Aguinaco, {\'I}nigo and Chrysostomou, Dimitris and B{\o}gh, Simon and Arana-Arexolaleiba, Nestor},
  booktitle={2023 IEEE/SICE International Symposium on System Integration (SII)},
  pages={1--5},
  year={2023},
  organization={IEEE}
}

@article{prata2023state,
  title={State of the art in robotic surgery with Hugo RAS system: feasibility, safety and clinical applications},
  author={Prata, Francesco and Ragusa, Alberto and Tempesta, Claudia and Iannuzzi, Andrea and Tedesco, Francesco and Cacciatore, Loris and Raso, Gianluigi and Civitella, Angelo and Tuzzolo, Piergiorgio and Call{\`e}, Pasquale and others},
  journal={Journal of personalized medicine},
  volume={13},
  number={8},
  pages={1233},
  year={2023},
  publisher={MDPI}
}

@article{celotto2024vinci,
  title={Da Vinci single-port robotic system current application and future perspective in general surgery: a scoping review},
  author={Celotto, Francesco and Ramacciotti, Niccol{\`o} and Mangano, Alberto and Danieli, Giacomo and Pinto, Federico and Lopez, Paula and Ducas, Alvaro and Cassiani, Jessica and Morelli, Luca and Spolverato, Gaya and others},
  journal={Surgical Endoscopy},
  volume={38},
  number={9},
  pages={4814--4830},
  year={2024},
  publisher={Springer}
}

@article{reed2022generalist,
  title={A generalist agent},
  author={Reed, Scott and Zolna, Konrad and Parisotto, Emilio and Colmenarejo, Sergio Gomez and Novikov, Alexander and Barth-Maron, Gabriel and Gimenez, Mai and Sulsky, Yury and Kay, Jackie and Springenberg, Jost Tobias and others},
  journal={arXiv preprint arXiv:2205.06175},
  year={2022}
}

@inproceedings{driess2023palm,
author = {Driess, Danny and Xia, Fei and Sajjadi, Mehdi S. M. and Lynch, Corey and Chowdhery, Aakanksha and Ichter, Brian and Wahid, Ayzaan and Tompson, Jonathan and Vuong, Quan and Yu, Tianhe and Huang, Wenlong and Chebotar, Yevgen and Sermanet, Pierre and Duckworth, Daniel and Levine, Sergey and Vanhoucke, Vincent and Hausman, Karol and Toussaint, Marc and Greff, Klaus and Zeng, Andy and Mordatch, Igor and Florence, Pete},
title = {PaLM-E: an embodied multimodal language model},
year = {2023},
publisher = {JMLR.org},
booktitle = {Proceedings of the 40th International Conference on Machine Learning},
articleno = {340},
numpages = {20},
location = {Honolulu, Hawaii, USA},
series = {ICML'23}
}

@article{tirinzoni2025zero,
  title={Zero-shot whole-body humanoid control via behavioral foundation models},
  author={Tirinzoni, Andrea and Touati, Ahmed and Farebrother, Jesse and Guzek, Mateusz and Kanervisto, Anssi and Xu, Yingchen and Lazaric, Alessandro and Pirotta, Matteo},
  journal={arXiv preprint arXiv:2504.11054},
  year={2025}
}

@inproceedings{chi2023diffusion,
	title={Diffusion Policy: Visuomotor Policy Learning via Action Diffusion},
	author={Chi, Cheng and Feng, Siyuan and Du, Yilun and Xu, Zhenjia and Cousineau, Eric and Burchfiel, Benjamin and Song, Shuran},
	booktitle={Proceedings of Robotics: Science and Systems (RSS)},
	year={2023}
}

@article{hu2023gaia,
  title={Gaia-1: A generative world model for autonomous driving},
  author={Hu, Anthony and Russell, Lloyd and Yeo, Hudson and Murez, Zak and Fedoseev, George and Kendall, Alex and Shotton, Jamie and Corrado, Gianluca},
  journal={arXiv preprint arXiv:2309.17080},
  year={2023}
}

@article{russell2025gaia,
  title={Gaia-2: A controllable multi-view generative world model for autonomous driving},
  author={Russell, Lloyd and Hu, Anthony and Bertoni, Lorenzo and Fedoseev, George and Shotton, Jamie and Arani, Elahe and Corrado, Gianluca},
  journal={arXiv preprint arXiv:2503.20523},
  year={2025}
}

@article{lan2023end,
  title={End-to-end planning of autonomous driving in industry and academia: 2022-2023},
  author={Lan, Gongjin and Hao, Qi},
  journal={arXiv preprint arXiv:2401.08658},
  year={2023}
}

@online{nvidia2025drive,
  author       = {Wu, Xinzhou},
  title        = {{NVIDIA DRIVE Full-Stack Autonomous Vehicle Software Rolls Out}},
  year         = {2025},
  month        = jun,
  day          = {11},
  note         = {NVIDIA GTC Paris announcement},
  url          = {https://blogs.nvidia.com/blog/drive-full-stack-av-software-europe/}
}

@article{hakami2024strategies,
  title={Strategies for overcoming data scarcity, imbalance, and feature selection challenges in machine learning models for predictive maintenance},
  author={Hakami, Ali},
  journal={Scientific Reports},
  volume={14},
  number={1},
  pages={9645},
  year={2024},
  publisher={Nature Publishing Group UK London}
}

@article{smuha2025regulation,
  title={Regulation 2024/1689 of the Eur. Parl. \& Council of June 13, 2024 (EU Artificial Intelligence Act)},
  author={Smuha, Nathalie A},
  journal={International Legal Materials},
  pages={1--148},
  year={2025},
  publisher={Cambridge University Press}
}

@article{hanna2025ethical,
  title={Ethical and bias considerations in artificial intelligence/machine learning},
  author={Hanna, Matthew G and Pantanowitz, Liron and Jackson, Brian and Palmer, Octavia and Visweswaran, Shyam and Pantanowitz, Joshua and Deebajah, Mustafa and Rashidi, Hooman H},
  journal={Modern Pathology},
  volume={38},
  number={3},
  pages={100686},
  year={2025},
  publisher={Elsevier}
}

@misc{razavi2025benchmarkingpromptsensitivitylarge,
      title={Benchmarking Prompt Sensitivity in Large Language Models}, 
      author={Amirhossein Razavi and Mina Soltangheis and Negar Arabzadeh and Sara Salamat and Morteza Zihayat and Ebrahim Bagheri},
      year={2025},
      eprint={2502.06065},
      archivePrefix={arXiv},
      primaryClass={cs.CL},
      url={https://arxiv.org/abs/2502.06065}, 
}

@misc{hughes2024openendednessessentialartificialsuperhuman,
      title={Open-Endedness is Essential for Artificial Superhuman Intelligence}, 
      author={Edward Hughes and Michael Dennis and Jack Parker-Holder and Feryal Behbahani and Aditi Mavalankar and Yuge Shi and Tom Schaul and Tim Rocktaschel},
      year={2024},
      eprint={2406.04268},
      archivePrefix={arXiv},
      primaryClass={cs.LG},
      url={https://arxiv.org/abs/2406.04268}, 
}

@inproceedings{wan2025generative,
  title={Generative AI in Embodied Systems: System-Level Analysis of Performance, Efficiency and Scalability},
  author={Wan, Zishen and Qian, Jiayi and Du, Yuhang and Jabbour, Jason and Du, Yilun and Zhao, Yang and Raychowdhury, Arijit and Krishna, Tushar and Reddi, Vijay Janapa},
  booktitle={2025 IEEE International Symposium on Performance Analysis of Systems and Software (ISPASS)},
  pages={26--37},
  year={2025},
  organization={IEEE}
}

@article{xing2025towards,
  title={Towards robust and secure embodied ai: A survey on vulnerabilities and attacks},
  author={Xing, Wenpeng and Li, Minghao and Li, Mohan and Han, Meng},
  journal={arXiv preprint arXiv:2502.13175},
  year={2025}
}

@article{schneider2024explainable,
  title={Explainable Generative AI (GenXAI): a survey, conceptualization, and research agenda},
  author={Schneider, Johannes},
  journal={Artificial Intelligence Review},
  volume={57},
  number={11},
  pages={289},
  year={2024},
  publisher={Springer}
}

@article{jegham2025hungry,
  title={How hungry is ai? benchmarking energy, water, and carbon footprint of llm inference},
  author={Jegham, Nidhal and Abdelatti, Marwan and Elmoubarki, Lassad and Hendawi, Abdeltawab},
  journal={arXiv preprint arXiv:2505.09598},
  year={2025}
}

@misc{Mucci2024,
  author       = {Tim Mucci},
  title        = {The future of {AI}: Trends shaping the next 10 years},
  year         = {2024},
  month        = {Oct 11},
  howpublished = {IBM Think website},
  url          = {https://www.ibm.com/think/insights/artificial-intelligence-future},
  note         = {Accessed: 2025-09-01}
}

@article{hu2023toward,
  title={Toward general-purpose robots via foundation models: A survey and meta-analysis},
  author={Hu, Yafei and Xie, Quanting and Jain, Vidhi and Francis, Jonathan and Patrikar, Jay and Keetha, Nikhil and Kim, Seungchan and Xie, Yaqi and Zhang, Tianyi and Fang, Hao-Shu and others},
  journal={arXiv preprint arXiv:2312.08782},
  year={2023}
}

@article{khan2025surgical,
  title={Surgical Scene Understanding in the Era of Foundation AI Models: A Comprehensive Review},
  author={Khan, Ufaq and Nawaz, Umair and Qayyum, Adnan and Ashraf, Shazad and Bilal, Muhammad and Qadir, Junaid},
  journal={arXiv preprint arXiv:2502.14886},
  year={2025}
}

@article{wang2024surgical,
  title={Surgical-lvlm: Learning to adapt large vision-language model for grounded visual question answering in robotic surgery},
  author={Wang, Guankun and Bai, Long and Nah, Wan Jun and Wang, Jie and Zhang, Zhaoxi and Chen, Zhen and Wu, Jinlin and Islam, Mobarakol and Liu, Hongbin and Ren, Hongliang},
  journal={arXiv preprint arXiv:2405.10948},
  year={2024}
}

@article{koh2018efficacy,
  title={Efficacy and safety of robotic procedures performed using the da Vinci robotic surgical system at a single institute in Korea: experience with 10000 cases},
  author={Koh, Dong Hoon and Jang, Won Sik and Park, Jae Won and Ham, Won Sik and Han, Woong Kyu and Rha, Koon Ho and Choi, Young Deuk},
  journal={Yonsei medical journal},
  volume={59},
  number={8},
  pages={975--981},
  year={2018},
  publisher={Yonsei University College of Medicine}
}

@article{nakajima2021cybernic,
  title={Cybernic treatment with wearable cyborg Hybrid Assistive Limb (HAL) improves ambulatory function in patients with slowly progressive rare neuromuscular diseases: a multicentre, randomised, controlled crossover trial for efficacy and safety (NCY-3001)},
  author={Nakajima, Takashi and Sankai, Yoshiyuki and Takata, Shinjiro and Kobayashi, Yoko and Ando, Yoshihito and Nakagawa, Masanori and Saito, Toshio and Saito, Kayoko and Ishida, Chiho and Tamaoka, Akira and others},
  journal={Orphanet journal of rare diseases},
  volume={16},
  number={1},
  pages={304},
  year={2021},
  publisher={Springer}
}

@article{sheng2025comprehensive,
  title={A Comprehensive Review of Humanoid Robots},
  author={Sheng, Qincheng and Zhou, Zhongxiang and Li, Jinhao and Mi, Xiangyu and Xiang, Pingyu and Chen, Zhenghan and Xu, Haocheng and Jia, Shenhan and Wu, Xiyang and Cui, Yuxiang and others},
  journal={SmartBot},
  volume={1},
  number={1},
  pages={e12008},
  year={2025},
  publisher={Wiley Online Library}
}

@article{bjorck2025gr00t,
  title={Gr00t n1: An open foundation model for generalist humanoid robots},
  author={Bjorck, Johan and Casta{\~n}eda, Fernando and Cherniadev, Nikita and Da, Xingye and Ding, Runyu and Fan, Linxi and Fang, Yu and Fox, Dieter and Hu, Fengyuan and Huang, Spencer and others},
  journal={arXiv preprint arXiv:2503.14734},
  year={2025}
}

@article{kuindersma2016optimization,
  title={Optimization-based locomotion planning, estimation, and control design for the atlas humanoid robot},
  author={Kuindersma, Scott and Deits, Robin and Fallon, Maurice and Valenzuela, Andr{\'e}s and Dai, Hongkai and Permenter, Frank and Koolen, Twan and Marion, Pat and Tedrake, Russ},
  journal={Autonomous robots},
  volume={40},
  number={3},
  pages={429--455},
  year={2016},
  publisher={Springer}
}

@misc{Dresser2023AmazonRobotics,
  author       = {Scott Dresser},
  title        = {Amazon announces 2 new ways it's using robots to assist employees and deliver for customers},
  howpublished = {About Amazon (Amazon Robotics blog)},
  year         = {2023},
  month        = oct,
}

@online{kalil2025tesla,
  author    = {Mike Kalil},
  title     = {Tesla Bot’s New Human-Level Learning Milestone},
  year      = {2025},
  url       = {https://mikekalil.com/blog/tesla-optimus-video-learning/},
  note      = {Accessed on September 1, 2025}
}

@misc{dehghan2023dynamicagvtaskallocation,
      title={Dynamic AGV Task Allocation in Intelligent Warehouses}, 
      author={Arash Dehghan and Mucahit Cevik and Merve Bodur},
      year={2023},
      eprint={2312.16026},
      archivePrefix={arXiv},
      primaryClass={math.OC},
      url={https://arxiv.org/abs/2312.16026}, 
}

@misc{greenawalt_robots_2025,
  author       = {Tyler Greenawalt},
  title        = {Amazon has more than 750,000 robots that sort, lift, and carry packages---see them in action},
  organization = {About Amazon},
  year         = {2025},
  month        = {June},
  day          = {11},
  howpublished = {\url{https://www.aboutamazon.com/news/operations/amazon-robotics-robots-fulfillment-center}},
  note         = {Accessed: 2025-09-02}
}

@misc{walmart_route_optimization_2024,
  organization = {Walmart Inc.},
  title        = {Walmart Commerce Technologies Launches AI-Powered Logistics Product},
  year         = {2024},
  month        = {March},
  day          = {14},
  howpublished = {\url{https://corporate.walmart.com/news/2024/03/14/walmart-commerce-technologies-launches-ai-powered-logistics-product}},
}

@misc{bmw2025virtualfactory,
  author       = {Moritz Schmerbeck},
  title        = {BMW Group scales Virtual Factory},
  howpublished = {BMW Group Press Release},
  year         = {2025},
  month        = {Jun 11},
  note         = {Press release, BMW Group PressClub},
  url          = {https://www.press.bmwgroup.com/global/article/detail/T0450699EN/bmw-group-scales-virtual-factory?showMedia=photo}
}

@misc{Graser2025GenAI4Q,
  author       = {Saskia Graser},
  title        = {Artificial intelligence as a quality booster},
  howpublished = {Press release, BMW Group PressClub Global},
  month        = apr,
  day          = {28},
  year         = {2025},
  note         = {Pilot project “GenAI4Q” at BMW Group Plant Regensburg enables tailored AI-based quality checks in vehicle assembly},
  url          = {https://www.press.bmwgroup.com/global/article/detail/T0449729EN/artificial-intelligence-as-a-quality-booster?language=en}
}

@article{o2021ai,
  title={AI-driven robotic laboratories show promise},
  author={O’Neill, Sean},
  journal={Engineering},
  volume={7},
  number={10},
  pages={1351},
  year={2021},
  publisher={Elsevier}
}

@article{m2024augmenting,
  title={Augmenting large language models with chemistry tools},
  author={M. Bran, Andres and Cox, Sam and Schilter, Oliver and Baldassari, Carlo and White, Andrew D and Schwaller, Philippe},
  journal={Nature Machine Intelligence},
  volume={6},
  number={5},
  pages={525--535},
  year={2024},
  publisher={Nature Publishing Group UK London}
}

@article{das2024overview,
  title={An overview on the role of artificial intelligence in modern advancements of material science},
  author={Das, Mayukh and Perez, Teresa Castillo and Shetty, Dasharathraj and Hiremath, Pavan and Naik, Nithesh and Bhat, Ritesh},
  journal={ES General},
  volume={5},
  pages={1183},
  year={2024},
  publisher={Engineered Science Publisher}
}

@article{liu2025generative,
  title={Generative physical ai in vision: A survey},
  author={Liu, Daochang and Zhang, Junyu and Dinh, Anh-Dung and Park, Eunbyung and Zhang, Shichao and Mian, Ajmal and Shah, Mubarak and Xu, Chang},
  journal={arXiv preprint arXiv:2501.10928},
  year={2025}
}

@article{shademan2016supervised,
  title={Supervised autonomous robotic soft tissue surgery},
  author={Shademan, Azad and Decker, Ryan S and Opfermann, Justin D and Leonard, Simon and Krieger, Axel and Kim, Peter CW},
  journal={Science translational medicine},
  volume={8},
  number={337},
  pages={337ra64--337ra64},
  year={2016},
  publisher={American Association for the Advancement of Science}
}

@article{lykov2024industry,
  title={Industry 6.0: New generation of industry driven by generative ai and swarm of heterogeneous robots},
  author={Lykov, Artem and Cabrera, Miguel Altamirano and Konenkov, Mikhail and Serpiva, Valerii and Gbagbe, Koffivi Fidele and Alabbas, Ali and Fedoseev, Aleksey and Moreno, Luis and Khan, Muhammad Haris and Guo, Ziang and others},
  journal={arXiv preprint arXiv:2409.10106},
  year={2024}
}

@article{li2023actuation,
  title={Actuation mechanisms and applications for soft robots: A comprehensive review},
  author={Li, Weidong and Hu, Diangang and Yang, Lei},
  journal={Applied Sciences},
  volume={13},
  number={16},
  pages={9255},
  year={2023},
  publisher={MDPI}
}

@article{sliwowski2025reassemble,
  title={Reassemble: A multimodal dataset for contact-rich robotic assembly and disassembly},
  author={Sliwowski, Daniel and Jadav, Shail and Stanovcic, Sergej and Orbik, Jedrzej and Heidersberger, Johannes and Lee, Dongheui},
  journal={arXiv preprint arXiv:2502.05086},
  year={2025}
}

@article{karnan2022socially,
  title={Socially compliant navigation dataset (scand): A large-scale dataset of demonstrations for social navigation},
  author={Karnan, Haresh and Nair, Anirudh and Xiao, Xuesu and Warnell, Garrett and Pirk, S{\"o}ren and Toshev, Alexander and Hart, Justin and Biswas, Joydeep and Stone, Peter},
  journal={IEEE Robotics and Automation Letters},
  volume={7},
  number={4},
  pages={11807--11814},
  year={2022},
  publisher={IEEE}
}

@article{muratore2022robot,
  title={Robot learning from randomized simulations: A review},
  author={Muratore, Fabio and Ramos, Fabio and Turk, Greg and Yu, Wenhao and Gienger, Michael and Peters, Jan},
  journal={Frontiers in Robotics and AI},
  volume={9},
  pages={799893},
  year={2022},
  publisher={Frontiers Media SA}
}

@inproceedings{kar2019meta,
  title={Meta-sim: Learning to generate synthetic datasets},
  author={Kar, Amlan and Prakash, Aayush and Liu, Ming-Yu and Cameracci, Eric and Yuan, Justin and Rusiniak, Matt and Acuna, David and Torralba, Antonio and Fidler, Sanja},
  booktitle={Proceedings of the IEEE/CVF international conference on computer vision},
  pages={4551--4560},
  year={2019}
}

@article{jiang2024robots,
  title={Robots pre-train robots: Manipulation-centric robotic representation from large-scale robot datasets},
  author={Jiang, Guangqi and Sun, Yifei and Huang, Tao and Li, Huanyu and Liang, Yongyuan and Xu, Huazhe},
  journal={arXiv preprint arXiv:2410.22325},
  year={2024}
}

@article{zhang2021reinforcement,
  title={Reinforcement learning for robot research: A comprehensive review and open issues},
  author={Zhang, Tengteng and Mo, Hongwei},
  journal={International Journal of Advanced Robotic Systems},
  volume={18},
  number={3},
  pages={17298814211007305},
  year={2021},
  publisher={SAGE Publications Sage UK: London, England}
}

@article{ibarz2021train,
  title={How to train your robot with deep reinforcement learning: lessons we have learned},
  author={Ibarz, Julian and Tan, Jie and Finn, Chelsea and Kalakrishnan, Mrinal and Pastor, Peter and Levine, Sergey},
  journal={The International Journal of Robotics Research},
  volume={40},
  number={4-5},
  pages={698--721},
  year={2021},
  publisher={SAGE Publications Sage UK: London, England}
}

@inproceedings{haarnoja2018soft,
  title={Soft actor-critic: Off-policy maximum entropy deep reinforcement learning with a stochastic actor},
  author={Haarnoja, Tuomas and Zhou, Aurick and Abbeel, Pieter and Levine, Sergey},
  booktitle={International conference on machine learning},
  pages={1861--1870},
  year={2018},
  organization={Pmlr}
}

@article{schulman2017proximal,
  title={Proximal policy optimization algorithms},
  author={Schulman, John and Wolski, Filip and Dhariwal, Prafulla and Radford, Alec and Klimov, Oleg},
  journal={arXiv preprint arXiv:1707.06347},
  year={2017}
}

@article{kulkarni2016hierarchical,
  title={Hierarchical deep reinforcement learning: Integrating temporal abstraction and intrinsic motivation},
  author={Kulkarni, Tejas D and Narasimhan, Karthik and Saeedi, Ardavan and Tenenbaum, Josh},
  journal={Advances in neural information processing systems},
  volume={29},
  year={2016}
}

@article{christiano2017deep,
  title={Deep reinforcement learning from human preferences},
  author={Christiano, Paul F and Leike, Jan and Brown, Tom and Martic, Miljan and Legg, Shane and Amodei, Dario},
  journal={Advances in neural information processing systems},
  volume={30},
  year={2017}
}

@article{argall2009survey,
  title={A survey of robot learning from demonstration},
  author={Argall, Brenna D and Chernova, Sonia and Veloso, Manuela and Browning, Brett},
  journal={Robotics and autonomous systems},
  volume={57},
  number={5},
  pages={469--483},
  year={2009},
  publisher={Elsevier}
}

@inproceedings{jha2022imitation,
  title={Imitation and supervised learning of compliance for robotic assembly},
  author={Jha, Devesh K and Romeres, Diego and Yerazunis, William and Nikovski, Daniel},
  booktitle={2022 European Control Conference (ECC)},
  pages={1882--1889},
  year={2022},
  organization={IEEE}
}

@article{bengio2013representation,
  title={Representation learning: A review and new perspectives},
  author={Bengio, Yoshua and Courville, Aaron and Vincent, Pascal},
  journal={IEEE transactions on pattern analysis and machine intelligence},
  volume={35},
  number={8},
  pages={1798--1828},
  year={2013},
  publisher={IEEE}
}

@inproceedings{pathak2017curiosity,
  title={Curiosity-driven exploration by self-supervised prediction},
  author={Pathak, Deepak and Agrawal, Pulkit and Efros, Alexei A and Darrell, Trevor},
  booktitle={International conference on machine learning},
  pages={2778--2787},
  year={2017},
  organization={PMLR}
}

@inproceedings{pinto2016supersizing,
  title={Supersizing self-supervision: Learning to grasp from 50k tries and 700 robot hours},
  author={Pinto, Lerrel and Gupta, Abhinav},
  booktitle={2016 IEEE international conference on robotics and automation (ICRA)},
  pages={3406--3413},
  year={2016},
  organization={IEEE}
}

@article{makoviychuk2021isaac,
  title={Isaac gym: High performance gpu-based physics simulation for robot learning},
  author={Makoviychuk, Viktor and Wawrzyniak, Lukasz and Guo, Yunrong and Lu, Michelle and Storey, Kier and Macklin, Miles and Hoeller, David and Rudin, Nikita and Allshire, Arthur and Handa, Ankur and others},
  journal={arXiv preprint arXiv:2108.10470},
  year={2021}
}

@inproceedings{todorov2012mujoco,
  title={Mujoco: A physics engine for model-based control},
  author={Todorov, Emanuel and Erez, Tom and Tassa, Yuval},
  booktitle={2012 IEEE/RSJ international conference on intelligent robots and systems},
  pages={5026--5033},
  year={2012},
  organization={IEEE}
}

@inproceedings{tobin2017domain,
  title={Domain randomization for transferring deep neural networks from simulation to the real world},
  author={Tobin, Josh and Fong, Rachel and Ray, Alex and Schneider, Jonas and Zaremba, Wojciech and Abbeel, Pieter},
  booktitle={2017 IEEE/RSJ international conference on intelligent robots and systems (IROS)},
  pages={23--30},
  year={2017},
  organization={IEEE}
}

@article{liu2023libero,
  title={Libero: Benchmarking knowledge transfer for lifelong robot learning},
  author={Liu, Bo and Zhu, Yifeng and Gao, Chongkai and Feng, Yihao and Liu, Qiang and Zhu, Yuke and Stone, Peter},
  journal={Advances in Neural Information Processing Systems},
  volume={36},
  pages={44776--44791},
  year={2023}
}

@article{james2020rlbench,
  title={Rlbench: The robot learning benchmark \& learning environment},
  author={James, Stephen and Ma, Zicong and Arrojo, David Rovick and Davison, Andrew J},
  journal={IEEE Robotics and Automation Letters},
  volume={5},
  number={2},
  pages={3019--3026},
  year={2020},
  publisher={IEEE}
}

@inproceedings{o2024open,
  title={Open x-embodiment: Robotic learning datasets and rt-x models: Open x-embodiment collaboration 0},
  author={O’Neill, Abby and Rehman, Abdul and Maddukuri, Abhiram and Gupta, Abhishek and Padalkar, Abhishek and Lee, Abraham and Pooley, Acorn and Gupta, Agrim and Mandlekar, Ajay and Jain, Ajinkya and others},
  booktitle={2024 IEEE International Conference on Robotics and Automation (ICRA)},
  pages={6892--6903},
  year={2024},
  organization={IEEE}
}

@article{lisondra2025embodied,
  title={Embodied AI with Foundation Models for Mobile Service Robots: A Systematic Review},
  author={Lisondra, Matthew and Benhabib, Beno and Nejat, Goldie},
  journal={arXiv preprint arXiv:2505.20503},
  year={2025}
}

@article{xiao2025robot,
  title={Robot learning in the era of foundation models: A survey},
  author={Xiao, Xuan and Liu, Jiahang and Wang, Zhipeng and Zhou, Yanmin and Qi, Yong and Jiang, Shuo and He, Bin and Cheng, Qian},
  journal={Neurocomputing},
  pages={129963},
  year={2025},
  publisher={Elsevier}
}

@article{xu2024survey,
  title={A survey on robotics with foundation models: toward embodied ai},
  author={Xu, Zhiyuan and Wu, Kun and Wen, Junjie and Li, Jinming and Liu, Ning and Che, Zhengping and Tang, Jian},
  journal={arXiv preprint arXiv:2402.02385},
  year={2024}
}

@article{zhang2025generative,
  title={Generative artificial intelligence in robotic manipulation: A survey},
  author={Zhang, Kun and Yun, Peng and Cen, Jun and Cai, Junhao and Zhu, Didi and Yuan, Hangjie and Zhao, Chao and Feng, Tao and Wang, Michael Yu and Chen, Qifeng and others},
  journal={arXiv preprint arXiv:2503.03464},
  year={2025}
}

@inproceedings{devlin2019bert,
  title={Bert: Pre-training of deep bidirectional transformers for language understanding},
  author={Devlin, Jacob and Chang, Ming-Wei and Lee, Kenton and Toutanova, Kristina},
  booktitle={Proceedings of the 2019 conference of the North American chapter of the association for computational linguistics: human language technologies, volume 1 (long and short papers)},
  pages={4171--4186},
  year={2019}
}

@article{back2025graspclutter6d,
  title={GraspClutter6D: A Large-scale Real-world Dataset for Robust Perception and Grasping in Cluttered Scenes},
  author={Back, Seunghyeok and Lee, Joosoon and Kim, Kangmin and Rho, Heeseon and Lee, Geonhyup and Kang, Raeyoung and Lee, Sangbeom and Noh, Sangjun and Lee, Youngjin and Lee, Taeyeop and others},
  journal={arXiv preprint arXiv:2504.06866},
  year={2025}
}

@article{o2025exploring,
  title={Exploring gpt-4 for robotic agent strategy with real-time state feedback and a reactive behaviour framework},
  author={O'Brien, Thomas and Sims, Ysobel},
  journal={arXiv preprint arXiv:2503.23601},
  year={2025}
}

@article{hao2025visual,
  title={Visual Large Language Models Exhibit Human-Level Cognitive Flexibility in the Wisconsin Card Sorting Test},
  author={Hao, Guangfu and Alexandre, Frederic and Yu, Shan},
  journal={arXiv preprint arXiv:2505.22112},
  year={2025}
}

@article{team2025gemini,
  title={Gemini robotics: Bringing ai into the physical world},
  author={Team, Gemini Robotics and Abeyruwan, Saminda and Ainslie, Joshua and Alayrac, Jean-Baptiste and Arenas, Montserrat Gonzalez and Armstrong, Travis and Balakrishna, Ashwin and Baruch, Robert and Bauza, Maria and Blokzijl, Michiel and others},
  journal={arXiv preprint arXiv:2503.20020},
  year={2025}
}

@article{khandelwal2023large,
  title={Large content and behavior models to understand, simulate, and optimize content and behavior},
  author={Khandelwal, Ashmit and Agrawal, Aditya and Bhattacharyya, Aanisha and Singla, Yaman K and Singh, Somesh and Bhattacharya, Uttaran and Dasgupta, Ishita and Petrangeli, Stefano and Shah, Rajiv Ratn and Chen, Changyou and others},
  journal={arXiv preprint arXiv:2309.00359},
  year={2023}
}

@article{team2024octo,
  title={Octo: An open-source generalist robot policy},
  author={Team, Octo Model and Ghosh, Dibya and Walke, Homer and Pertsch, Karl and Black, Kevin and Mees, Oier and Dasari, Sudeep and Hejna, Joey and Kreiman, Tobias and Xu, Charles and others},
  journal={arXiv preprint arXiv:2405.12213},
  year={2024}
}

@article{qin2024worldsimbench,
  title={Worldsimbench: Towards video generation models as world simulators},
  author={Qin, Yiran and Shi, Zhelun and Yu, Jiwen and Wang, Xijun and Zhou, Enshen and Li, Lijun and Yin, Zhenfei and Liu, Xihui and Sheng, Lu and Shao, Jing and others},
  journal={arXiv preprint arXiv:2410.18072},
  year={2024}
}

@article{li2025worldmodelbench,
  title={Worldmodelbench: Judging video generation models as world models},
  author={Li, Dacheng and Fang, Yunhao and Chen, Yukang and Yang, Shuo and Cao, Shiyi and Wong, Justin and Luo, Michael and Wang, Xiaolong and Yin, Hongxu and Gonzalez, Joseph E and others},
  journal={arXiv preprint arXiv:2502.20694},
  year={2025}
}

@inproceedings{dasari2025ingredients,
  title={The ingredients for robotic diffusion transformers},
  author={Dasari, Sudeep and Mees, Oier and Zhao, Sebastian and Srirama, Mohan Kumar and Levine, Sergey},
  booktitle={2025 IEEE International Conference on Robotics and Automation (ICRA)},
  pages={15617--15625},
  year={2025},
  organization={IEEE}
}

@article{mandlekar2021matters,
  title={What matters in learning from offline human demonstrations for robot manipulation},
  author={Mandlekar, Ajay and Xu, Danfei and Wong, Josiah and Nasiriany, Soroush and Wang, Chen and Kulkarni, Rohun and Fei-Fei, Li and Savarese, Silvio and Zhu, Yuke and Mart{\'\i}n-Mart{\'\i}n, Roberto},
  journal={arXiv preprint arXiv:2108.03298},
  year={2021}
}

@article{mees2022calvin,
  title={Calvin: A benchmark for language-conditioned policy learning for long-horizon robot manipulation tasks},
  author={Mees, Oier and Hermann, Lukas and Rosete-Beas, Erick and Burgard, Wolfram},
  journal={IEEE Robotics and Automation Letters},
  volume={7},
  number={3},
  pages={7327--7334},
  year={2022},
  publisher={IEEE}
}

@article{hou2024diffusion,
  title={Diffusion transformer policy},
  author={Hou, Zhi and Zhang, Tianyi and Xiong, Yuwen and Pu, Hengjun and Zhao, Chengyang and Tong, Ronglei and Qiao, Yu and Dai, Jifeng and Chen, Yuntao},
  journal={arXiv preprint arXiv:2410.15959},
  year={2024}
}

@article{duan2025worldscore,
  title={Worldscore: A unified evaluation benchmark for world generation},
  author={Duan, Haoyi and Yu, Hong-Xing and Chen, Sirui and Fei-Fei, Li and Wu, Jiajun},
  journal={arXiv preprint arXiv:2504.00983},
  year={2025}
}

@article{long2025survey,
  title={A Survey: Learning Embodied Intelligence from Physical Simulators and World Models},
  author={Long, Xiaoxiao and Zhao, Qingrui and Zhang, Kaiwen and Zhang, Zihao and Wang, Dingrui and Liu, Yumeng and Shu, Zhengjie and Lu, Yi and Wang, Shouzheng and Wei, Xinzhe and others},
  journal={arXiv preprint arXiv:2507.00917},
  year={2025}
}

@inproceedings{mower2023ros,
  title={ROS-PyBullet Interface: A framework for reliable contact simulation and human-robot interaction},
  author={Mower, Christopher and Stouraitis, Theodoros and Moura, Joao and Rauch, Christian and Yan, Lei and Behabadi, Nazanin Zamani and Gienger, Michael and Vercauteren, Tom and Bergeles, Christos and Vijayakumar, Sethu},
  booktitle={Conference on robot learning},
  pages={1411--1423},
  year={2023},
  organization={PMLR}
}

@article{wang2025unified,
  title={Unified Vision-Language-Action Model},
  author={Wang, Yuqi and Li, Xinghang and Wang, Wenxuan and Zhang, Junbo and Li, Yingyan and Chen, Yuntao and Wang, Xinlong and Zhang, Zhaoxiang},
  journal={arXiv preprint arXiv:2506.19850},
  year={2025}
}

@article{Deloitte_article,
  author  = "{Deloitte}",
  title   = "Robotics and Physical AI: Intelligence in Motion",
  journal = "The Wall Street Journal",
  year    = "2025",
  month   = "November",
  day     = "11",
  url     = "https://deloitte.wsj.com/cio/robotics-and-physical-ai-intelligence-in-motion-3f6c000d"
}

@article{Forbes_JK_article,
  author  = "Kelly, Jack",
  title   = "The Rise Of AI-Powered Robotics, And The Future Of Work",
  journal = "Forbes",
  year    = "2025",
  month   = "April",
  day     = "15",
  url     = "https://www.forbes.com/sites/jackkelly/2025/04/15/the-rise-of-ai-powered-robotics-and-the-future-of-work/"
}

@article{GoldmanSachs_report,
  author  = "{Goldman Sachs}",
  title   = "The global market for humanoid robots could reach \$38 billion by 2035",
  journal = "Goldman Sachs Insights",
  year    = "2025",
  url     = "https://www.goldmansachs.com/insights/articles/the-global-market-for-robots-could-reach-38-billion-by-2035.html"
}

@techreport{Citi_report,
  author    = "{Citigroup}",
  title     = "The Rise of AI Robots",
  institution = "Citi GPS",
  year      = "2025",
  url       = "https://www.citigroup.com/global/insights/the-rise-of-ai-robots"
}

@article{muller2022self,
  title={Self-improving models for the intelligent digital twin: Towards closing the reality-to-simulation gap},
  author={M{\"u}ller, Manuel S and Jazdi, Nasser and Weyrich, Michael},
  journal={Ifac-Papersonline},
  volume={55},
  number={2},
  pages={126--131},
  year={2022},
  publisher={Elsevier}
}

@article{sun2024comprehensive,
  title={A comprehensive survey on embodied intelligence: Advancements, challenges, and future perspectives},
  author={Sun, Fuchun and Chen, Runfa and Ji, Tianying and Luo, Yu and Zhou, Huaidong and Liu, Huaping},
  journal={CAAI Artificial Intelligence Research},
  volume={3},
  pages={9150042},
  year={2024},
  publisher={清华大学出版社}
}

@article{mon2025embodied,
  title={Embodied large language models enable robots to complete complex tasks in unpredictable environments},
  author={Mon-Williams, Ruaridh and Li, Gen and Long, Ran and Du, Wenqian and Lucas, Christopher G},
  journal={Nature Machine Intelligence},
  pages={1--10},
  year={2025},
  publisher={Nature Publishing Group UK London}
}

@article{motta2023framework,
  title={A framework for FAIR robotic datasets},
  author={Motta, Corrado and Aracri, Simona and Ferretti, Roberta and Bibuli, Marco and Bruzzone, Gabriele and Caccia, Massimo and Odetti, Angelo and Ferreira, Fausto and de Pascalis, Francesca},
  journal={Scientific data},
  volume={10},
  number={1},
  pages={620},
  year={2023},
  publisher={Nature Publishing Group UK London}
}

@inproceedings{perez2018film,
  title={Film: Visual reasoning with a general conditioning layer},
  author={Perez, Ethan and Strub, Florian and De Vries, Harm and Dumoulin, Vincent and Courville, Aaron},
  booktitle={Proceedings of the AAAI conference on artificial intelligence},
  volume={32},
  number={1},
  year={2018}
}

@inproceedings{tan2019efficientnet,
  title={Efficientnet: Rethinking model scaling for convolutional neural networks},
  author={Tan, Mingxing and Le, Quoc},
  booktitle={International conference on machine learning},
  pages={6105--6114},
  year={2019},
  organization={PMLR}
}

@article{cer2018universal,
  title={Universal sentence encoder},
  author={Cer, Daniel and Yang, Yinfei and Kong, Sheng-yi and Hua, Nan and Limtiaco, Nicole and John, Rhomni St and Constant, Noah and Guajardo-Cespedes, Mario and Yuan, Steve and Tar, Chris and others},
  journal={arXiv preprint arXiv:1803.11175},
  year={2018}
}

@article{li2023model,
  title={Model compression for deep neural networks: A survey},
  author={Li, Zhuo and Li, Hengyi and Meng, Lin},
  journal={Computers},
  volume={12},
  number={3},
  pages={60},
  year={2023},
  publisher={MDPI}
}

@article{jouini2024survey,
  title={A survey of machine learning in edge computing: Techniques, frameworks, applications, issues, and research directions},
  author={Jouini, Oumayma and Sethom, Kaouthar and Namoun, Abdallah and Aljohani, Nasser and Alanazi, Meshari Huwaytim and Alanazi, Mohammad N},
  journal={Technologies},
  volume={12},
  number={6},
  pages={81},
  year={2024},
  publisher={MDPI}
}

@inproceedings{jeong2020selfsupervised,
  title        = {Self-Supervised Sim-to-Real Adaptation for Visual Robotic Manipulation},
  author       = {Jeong, Rae and Aytar, Yusuf and Khosid, David and Zhou, Yuxiang and Kay, Jackie and Lampe, Thomas and Bousmalis, Konstantinos and Nori, Francesco},
  booktitle    = {2020 IEEE International Conference on Robotics and Automation (ICRA)},
  year         = {2020},
  pages        = {2718--2724},
  publisher    = {IEEE},
  doi          = {10.1109/ICRA40945.2020.9197326}
}

@article{tang2022survey,
  title={Survey on digital twin edge networks (DITEN) toward 6G},
  author={Tang, Fengxiao and Chen, Xuehan and Rodrigues, Tiago Koketsu and Zhao, Ming and Kato, Nei},
  journal={IEEE Open Journal of the Communications Society},
  volume={3},
  pages={1360--1381},
  year={2022},
  publisher={IEEE}
}

@article{rodrigues2019machine,
  title={Machine learning meets computation and communication control in evolving edge and cloud: Challenges and future perspective},
  author={Rodrigues, Tiago Koketsu and Suto, Katsuya and Nishiyama, Hiroki and Liu, Jiajia and Kato, Nei},
  journal={IEEE Communications Surveys \& Tutorials},
  volume={22},
  number={1},
  pages={38--67},
  year={2019},
  publisher={IEEE}
}

@article{chen2024aigc,
  title={AIGC-based evolvable digital twin networks: A road to the intelligent metaverse},
  author={Chen, Xuehan and Luo, Linfeng and Tang, Fengxiao and Zhao, Ming and Kato, Nei},
  journal={IEEE Network},
  volume={38},
  number={6},
  pages={370--379},
  year={2024},
  publisher={IEEE}
}

@article{zhang2025review,
  title={A review of embodied intelligence systems: a three-layer framework integrating multimodal perception, world modeling, and structured strategies},
  author={Zhang, Yunwei and Tian, Jing and Xiong, Qiaochu},
  journal={Frontiers in Robotics and AI},
  volume={12},
  pages={1668910},
  year={2025},
  publisher={Frontiers}
}

@inproceedings{florence2022implicit,
  title={Implicit behavioral cloning},
  author={Florence, Pete and Lynch, Corey and Zeng, Andy and Ramirez, Oscar A and Wahid, Ayzaan and Downs, Laura and Wong, Adrian and Lee, Johnny and Mordatch, Igor and Tompson, Jonathan},
  booktitle={Conference on robot learning},
  pages={158--168},
  year={2022},
  organization={PMLR}
}

@article{ho2022video,
  title={Video diffusion models},
  author={Ho, Jonathan and Salimans, Tim and Gritsenko, Alexey and Chan, William and Norouzi, Mohammad and Fleet, David J},
  journal={Advances in neural information processing systems},
  volume={35},
  pages={8633--8646},
  year={2022}
}

@article{dihan2024digital,
  title={Digital twin: Data exploration, architecture, implementation and future},
  author={Dihan, Md Shezad and Akash, Anwar Islam and Tasneem, Zinat and Das, Prangon and Das, Sajal Kumar and Islam, Md Robiul and Islam, Md Manirul and Badal, Faisal R and Ali, Md Firoj and Ahamed, Md Hafiz and others},
  journal={Heliyon},
  volume={10},
  number={5},
  year={2024},
  publisher={Elsevier}
}

@inproceedings{heiden2021neuralsim,
  title={NeuralSim: Augmenting differentiable simulators with neural networks},
  author={Heiden, Eric and Millard, David and Coumans, Erwin and Sheng, Yizhou and Sukhatme, Gaurav S},
  booktitle={2021 IEEE International Conference on Robotics and Automation (ICRA)},
  pages={9474--9481},
  year={2021},
  organization={IEEE}
}

@inproceedings{ensinger2024learning,
  title={Learning hybrid dynamics models with simulator-informed latent states},
  author={Ensinger, Katharina and Ziesche, Sebastian and Trimpe, Sebastian},
  booktitle={Proceedings of the AAAI Conference on Artificial Intelligence},
  volume={38},
  number={11},
  pages={11892--11900},
  year={2024}
}

@article{lee2023rlaif,
  title={Rlaif: Scaling reinforcement learning from human feedback with ai feedback},
  author={Lee, Harrison and Phatale, Samrat and Mansoor, Hassan and Lu, Kellie Ren and Mesnard, Thomas and Ferret, Johan and Bishop, Colton and Hall, Ethan and Carbune, Victor and Rastogi, Abhinav},
  year={2023}
}

@article{li2025pin,
  title={Pin-wm: Learning physics-informed world models for non-prehensile manipulation},
  author={Li, Wenxuan and Zhao, Hang and Yu, Zhiyuan and Du, Yu and Zou, Qin and Hu, Ruizhen and Xu, Kai},
  journal={arXiv preprint arXiv:2504.16693},
  year={2025}
}

@article{leng2025physics,
  title={Physics-informed machine learning in intelligent manufacturing: a review},
  author={Leng, Jiewu and Zuo, Kaiwen and Xu, Caiyu and Zhou, Xueliang and Zheng, Shuai and Kang, Jiawen and Liu, Qiang and Chen, Xin and Shen, Weiming and Wang, Lihui and others},
  journal={Journal of Intelligent Manufacturing},
  pages={1--43},
  year={2025},
  publisher={Springer}
}

@article{leng2025diffusion,
  title={Diffusion model-driven smart design and manufacturing: Prospects and challenges},
  author={Leng, Jiewu and Su, Xuyang and Liu, Zean and Zhou, Lianhong and Chen, Chong and Guo, Xin and Wang, Yiwei and Wang, Ru and Zhang, Chao and Liu, Qiang and others},
  journal={Journal of manufacturing systems},
  volume={82},
  pages={561--577},
  year={2025},
  publisher={Elsevier}
}

@article{rovzanec2023human,
  title={Human-centric artificial intelligence architecture for industry 5.0 applications},
  author={Ro{\v{z}}anec, Jo{\v{z}}e M and Novalija, Inna and Zajec, Patrik and Kenda, Klemen and Tavakoli Ghinani, Hooman and Suh, Sungho and Bian, Sizhen and Veliou, Entso and Papamartzivanos, Dimitrios and Giannetsos, Thanassis and others},
  journal={International journal of production research},
  volume={61},
  number={20},
  pages={6847--6872},
  year={2023},
  publisher={Taylor \& Francis}
}

@article{leng2026aigc,
  title={AIGC-empowered smart manufacturing: Prospects and challenges},
  author={Leng, Jiewu and Zheng, Keyou and Li, Rongjie and Chen, Chong and Wang, Baicun and Liu, Qiang and Chen, Xin and Shen, Weiming},
  journal={Robotics and Computer-Integrated Manufacturing},
  volume={97},
  pages={103076},
  year={2026},
  publisher={Elsevier}
}

@article{sedlacek2025realm,
  title={REALM: A Real-to-Sim Validated Benchmark for Generalization in Robotic Manipulation},
  author={Sedlacek, Martin and Yefanov, Pavlo and Ponimatkin, Georgy and Bardhan, Jai and Pilc, Simon and Fourmy, Mederic and Kazakos, Evangelos and Snoek, Cees GM and Sivic, Josef and Petrik, Vladimir},
  journal={arXiv preprint arXiv:2512.19562},
  year={2025}
}

@article{tang2024defining,
  title={Defining and evaluating physical safety for large language models},
  author={Tang, Yung-Chen and Chen, Pin-Yu and Ho, Tsung-Yi},
  journal={arXiv preprint arXiv:2411.02317},
  year={2024}
}

@inproceedings{taylor2020learning,
  title={Learning for safety-critical control with control barrier functions},
  author={Taylor, Andrew and Singletary, Andrew and Yue, Yisong and Ames, Aaron},
  booktitle={Learning for dynamics and control},
  pages={708--717},
  year={2020},
  organization={PMLR}
}

@article{almujally2024multi,
  title={Multi-modal remote perception learning for object sensory data},
  author={Almujally, Nouf Abdullah and Rafique, Adnan Ahmed and Al Mudawi, Naif and Alazeb, Abdulwahab and Alonazi, Mohammed and Algarni, Asaad and Jalal, Ahmad and Liu, Hui},
  journal={Frontiers in Neurorobotics},
  volume={18},
  pages={1427786},
  year={2024},
  publisher={Frontiers Media SA}
}

@article{yuan2024transformer,
  title={Transformer in reinforcement learning for decision-making: a survey},
  author={Yuan, Weilin and Chen, Jiaxing and Chen, Shaofei and Feng, Dawei and Hu, Zhenzhen and Li, Peng and Zhao, Weiwei},
  journal={Frontiers of Information Technology \& Electronic Engineering},
  volume={25},
  number={6},
  pages={763--790},
  year={2024},
  publisher={ZUP}
}

@article{he2025neurodynamics,
  title={Neurodynamics-based visual servo predictive control for improving smooth movement of logistics omnidirectional robots},
  author={He, Defeng and Lin, Yegui and Dai, Zhijian and Yang, Simon X},
  journal={IEEE Transactions on Industrial Electronics},
  year={2025},
  publisher={IEEE}
}

@article{yang2025automated,
  title={Automated path-planning strategy for robotic inspection of underground utilities based on building information model},
  author={Yang, Zihan and Shu, Jiangpeng and Jiang, Jishuang and Han, Wentao and Wang, Yichang and Zhao, Liang and Bai, Yong},
  journal={Computer-Aided Civil and Infrastructure Engineering},
  volume={40},
  number={29},
  pages={5554--5575},
  year={2025},
  publisher={Wiley Online Library}
}

@article{yang2025global,
  title={The global industrial robot trade network: evolution and China’s rising international competitiveness},
  author={Yang, Huijie and Wei, Shaobin and Zhou, Haiyan and Hu, Feng and Chen, Yufeng and Hu, Hao},
  journal={Systems},
  volume={13},
  number={5},
  pages={361},
  year={2025},
  publisher={MDPI}
}

@article{sai2024generative,
  title={Generative AI for Industry 5.0: Analyzing the impact of ChatGPT, DALLE, and other models},
  author={Sai, Siva and Sai, Revant and Chamola, Vinay},
  journal={IEEE Open Journal of the Communications Society},
  volume={6},
  pages={3056--3066},
  year={2024},
  publisher={IEEE}
}

@article{shen2024evolutionary,
  title={Evolutionary computation-based self-supervised learning for image processing: a big data-driven approach to feature extraction and fusion for multispectral object detection},
  author={Shen, Xiaoyang and Li, Haibin and Shankar, Achyut and Viriyasitavat, Wattana and Chamola, Vinay},
  journal={Journal of Big Data},
  volume={11},
  number={1},
  pages={130},
  year={2024},
  publisher={Springer}
}

@article{kumar2025peeping,
  title={Peeping into the future: understanding and combating generative AI-based fake news},
  author={Kumar, Sanjeev and Sai, Siva and Chamola, Vinay and Gaur, Aanchal and Agarwal, Chitwan and Huang, Kaizhu and Hussain, Amir},
  journal={Cognitive Computation},
  volume={17},
  number={3},
  pages={103},
  year={2025},
  publisher={Springer}
}

@article{sai2024pivotal,
  title={Pivotal role of digital twins in the metaverse: A review},
  author={Sai, Siva and Sharma, Pulkit and Gaur, Aanchal and Chamola, Vinay},
  journal={Digital Communications and Networks},
  year={2024},
  publisher={Elsevier}
}

@article{sai2024device,
  title={On-device generative AI: the need, architectures, and challenges},
  author={Sai, Siva and Prasad, Manish and Dashore, Garima and Chamola, Vinay and Sikdar, Biplab},
  journal={IEEE Consumer Electronics Magazine},
  volume={14},
  number={4},
  pages={21--32},
  year={2024},
  publisher={IEEE}
}


\end{document}